\documentclass{article} % For LaTeX2e
\usepackage{iclr2026_conference,times}

\usepackage{amsmath,amsfonts,bm}

\def\eqref#1{equation~\ref{#1}}
\def\1{\bm{1}}

\DeclareMathAlphabet{\mathsfit}{\encodingdefault}{\sfdefault}{m}{sl}
\SetMathAlphabet{\mathsfit}{bold}{\encodingdefault}{\sfdefault}{bx}{n}

\usepackage{hyperref}
\usepackage{url}
\usepackage{pifont}
\usepackage{graphicx}
\usepackage{hyperref}       % hyperlinks
\usepackage{url}            % simple URL typesetting
\usepackage{booktabs}       % professional-quality tables

\usepackage{amsfonts}       % blackboard math symbols
\usepackage{xspace}

\usepackage{nicefrac}       % compact symbols for 1/2, etc.
\usepackage{microtype}      % microtypography
\usepackage{graphicx}
\usepackage{amsmath}
\usepackage{algorithm}      % 提供浮动体环境（如 algorithm 环境）
\usepackage{algorithmicx}   % 提供语法命令（推荐替代旧的 algorithmic）
\usepackage{algpseudocode}  % 提供常用伪代码关键字（如 If, While 等）
\usepackage{booktabs}
\usepackage{siunitx}
\usepackage{caption}
\usepackage{xcolor}
\usepackage[table]{xcolor} % 启用带颜色的表格
\usepackage{booktabs}      % 专业的三线表
\usepackage{siunitx}       % 精确的小数点对齐
\usepackage{caption}       % 改善标题样式
\usepackage{graphicx}      % 用于旋转符号等 (可选)
\usepackage{booktabs}
\usepackage{adjustbox}
\usepackage[table]{xcolor}
\usepackage{multirow}
\usepackage{adjustbox}
\usepackage{wrapfig}
\usepackage{textcomp}
\usepackage{float}
\usepackage{subfig}
\usepackage[table]{xcolor} % 用于颜色支持
\usepackage{siunitx}      % 用于对齐数字和单位
\usepackage{booktabs}     % 用于漂亮的表格线条 (toprule, midrule, bottomrule)
\usepackage{caption}      % 用于表格标题
\usepackage{siunitx}
\usepackage{multirow} 
\usepackage{makecell}
\usepackage{graphicx}
\usepackage{subcaption}
\usepackage{fvextra}
\usepackage{fancyvrb}
\usepackage{subcaption}
\usepackage[most]{tcolorbox}

\usepackage{fontawesome5}
\usepackage{pifont}
\usepackage{makecell}
\usepackage[most]{tcolorbox}
\usepackage{mdframed}
\title{Ego to World: Collaborative Spatial Reasoning in Embodied Systems via Reinforcement Learning}

\iclrfinalcopy
\author{
Heng Zhou$^{1,2,*}$ \quad
Li Kang$^{2,3,*}$ \quad
Yiran Qin$^{2,\ddagger}$ \quad
Xiufeng Song$^{2,3}$ \quad
Ao Yu$^{1}$ \quad \\
\
\textbf{Zilu Zhang}$^4$ \quad
\textbf{Haoming Song}$^{2,3}$ \quad
\textbf{Kaixin Xu}$^{2,5}$ \quad
\textbf{Yuchen Fan}$^{2,3}$ \quad \\
\
\textbf{Dongzhan Zhou}$^2$ \quad
\textbf{Xiaohong Liu}$^3$ \quad
\textbf{Ruimao Zhang}$^6$ \quad
\textbf{Philip Torr}$^7$ \quad \\
\
\textbf{Lei Bai}$^{2,\dagger}$ \quad
\textbf{Zhenfei Yin}$^{7,\dagger}$ \quad
\\
$^1$ University of Science and Technology of China \quad $^2$ Shanghai AI Laboratory \quad \\
$^3$ Shanghai Jiao Tong University \quad
$^4$ Beijing University of Posts and Telecommunications \quad \\
$^5$ Fudan University \quad
$^6$ Sun Yat-sen University \quad
$^7$ University of Oxford \quad \\
$^*$Equal contributions~~ $^\ddagger$Project leader~~ $^\dagger$Corresponding author \vspace{1mm} \\
\ \faEnvelope\ \texttt{hengzzzhou@gmail.com} \quad \faGithub\ \href{https://github.com/hengzzzhou/CoRL}{\texttt{Code}} \quad \faDatabase\ \href{https://huggingface.co/collections/henggg/corl}{\texttt{HuggingFace}} \quad \faGlobe\ \href{https://hengzzzhou.github.io/CoRL}{\texttt{Project Page}}\\
}

\begin{document}
\maketitle
\begin{abstract}
    Understanding the world from distributed, partial viewpoints is a fundamental challenge for embodied multi-agent systems. Each agent perceives the environment through an ego-centric view that is often limited by occlusion and ambiguity. To study this problem, we introduce the Ego-to-World (E2W) benchmark, which evaluates vision–language model’s ability to fuse heterogeneous viewpoints across three tasks: (i) global counting, (ii) relational location reasoning, and (iii) action-oriented grasping that requires predicting view-specific image coordinates.
To address this setting, we propose CoRL, a two-stage framework that combines Chain-of-Thought supervised fine-tuning with reinforcement learning using Group-Relative Policy Optimization. Its core component, the Cross-View Spatial Reward (CVSR), provides dense task-aligned feedback by linking reasoning steps to visual evidence, ensuring coherent cross-view entity resolution, and guiding the model toward correct final predictions.
Experiments on E2W show that CoRL consistently surpasses strong proprietary and open-source baselines on both reasoning and perception-grounding metrics, while ablations further confirm the necessity of each CVSR component. Beyond that, CoRL generalizes to external spatial reasoning benchmarks and enables effective real-world multi-robot manipulation with calibrated multi-camera rigs, demonstrating cross-view localization and successful grasp-and-place execution.
Together, E2W and CoRL provide a principled foundation for learning world-centric scene understanding from distributed, ego-centric observations, advancing collaborative embodied AI.
\end{abstract}

\section{Introduction}
    \label{sec:intro}

Recent advances in Vision--Language Models (VLMs) have catalyzed significant progress in embodied intelligence. By grounding natural language within visual perception, VLMs have enabled a diverse range of embodied tasks, from instruction following to interactive manipulation. This progress has spurred extensive research into VLM applications across domains such as robotics~\citep{kang2025viki}, navigation~\citep{wang2025rad}, and spatial reasoning~\citep{zhou2025roborefer,yin2025spatialmentalmodelinglimited}. Nevertheless, the majority of existing methodologies are confined to single-view scenarios, where perception is limited to an ego-centric or a fixed global viewpoint. Such a constraint inherently leads to incomplete scene understanding and restricted reasoning capabilities.

\begin{figure}[t]
    \centering
    \includegraphics[width=1.0\linewidth]{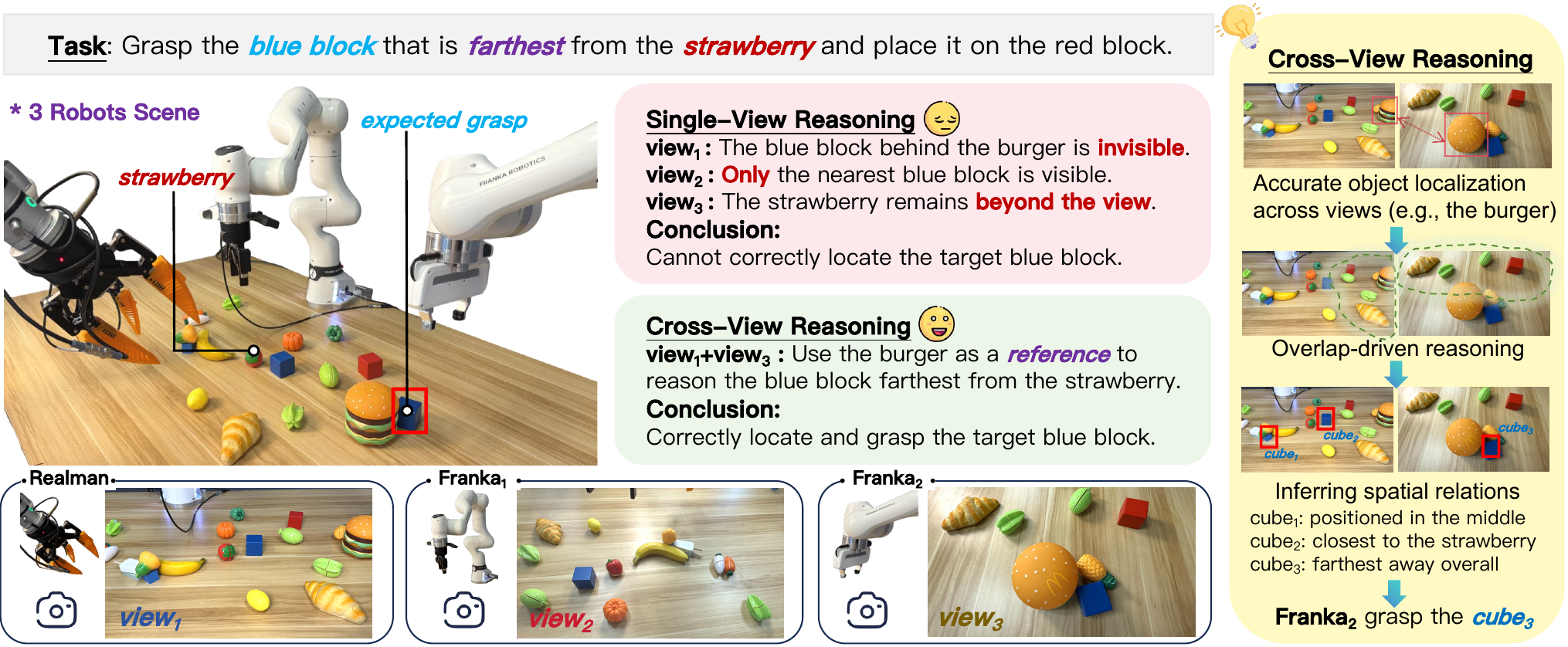}
    \caption{An illustration of collaborative spatial reasoning in embodied systems. Reasoning from a single viewpoint fails due to occlusions or a limited field of view. In contrast, cross-view compositional reasoning integrates multiple perspectives to correctly localize and grasp the target object---the blue block farthest from the strawberry.}
    \label{fig:crossview}
\end{figure}

In many real-world applications, multiple heterogeneous agents, such as cooperative service robots in domestic environments~\citep{10.1145/1121241.1121286} or Vehicle-to-Everything (V2X) systems in autonomous driving, operate concurrently~\citep{liu2023vehicletoeverythingautonomousdrivingsurvey}. Throughout this work, ``multi-agent'' refers to physically separate robots, each equipped with its own ego-centric sensor and limited field of view, rather than to multi-agent planning or decision-making. In these settings, multi-agent coordination is not merely beneficial but essential. A solitary viewpoint is fundamentally susceptible to occlusions and partial observations, whereas integrating complementary perspectives from multiple agents can provide richer contextual understanding. As illustrated in Figure~\ref{fig:crossview}, cross-view compositional reasoning empowers agents to surmount these limitations and execute spatially grounded actions with high fidelity.

From a practical standpoint, relying on a single global camera is often infeasible: global cameras require careful placement, suffer from occlusions, and cannot flexibly cover diverse interaction zones. In contrast, ego-centric sensors naturally accompany each agent, providing \emph{dynamic}, \emph{task-relevant}, and \emph{complementary} observations that yield richer spatial coverage. From a reasoning perspective, multi-view information is more expressive than any single global shot: while a global viewpoint may miss critical details due to perspective distortion or object self-occlusion, aggregating multiple close-range ego-centric views reduces ambiguity and significantly improves spatial grounding.

However, achieving such reasoning capabilities is profoundly challenging, as it necessitates the integration of heterogeneous viewpoints, the resolution of cross-view ambiguities, and the alignment of overlapping observations to construct a coherent scene representation.

To bridge this critical gap, we investigate the problem of collaborative spatial reasoning in multi-agent embodied systems. We formalize a novel setting wherein each agent contributes its partial, ego-centric observations, and the VLM must integrate these disparate inputs to perform complex spatial reasoning and perception tasks. To this end, we introduce the \textbf{Co}llaborative Spatial \textbf{R}easoning Reinforcement \textbf{L}earning (CoRL) framework, which is augmented by a novel \textbf{C}ross-\textbf{V}iew \textbf{S}patial \textbf{R}eward (CVSR). The CVSR is designed to explicitly incentivize policies that: (i) consistently localize objects across different viewpoints, (ii) effectively fuse complementary ego-centric information, and (iii) maximize collective scene coverage. By shaping the learning signal around these core principles of spatial grounding, CVSR compels the VLM to transcend single-view perception and develop robust collaborative spatial reasoning abilities. Through systematic comparisons of alternative training paradigms, we demonstrate that our proposed Supervised Fine-Tuning (SFT) followed by Reinforcement Learning (RL) pipeline, empowered by CVSR, consistently achieves state-of-the-art performance on collaborative perception tasks.

In summary, our contributions are threefold:  

$\diamond$ We pioneer the formalization of collaborative perception and reasoning with VLMs in multi-agent embodied systems, and introduce the E2W benchmark specifically tailored for this setting.

$\diamond$ We propose the CoRL framework, featuring the novel Cross-View Spatial Reward (CVSR), which explicitly incentivizes the fusion of ego-centric views and enhances spatial grounding.

$\diamond$ We empirically validate the effectiveness and generalizability of our approach , consistently showing substantial improvements over strong baselines.

\section{Related Work}
    \label{sec:related}

\paragraph{Embodied Multi-Agent Cooperation.}
Coordination among multiple agents is a fundamental challenge in embodied AI. Research in this domain has historically focused on high-level task allocation~\cite{obata2024lip,wang2024dart,liu2025language} and joint decision-making~\cite{zhang2023building,wang2025multi}. More recently, the advent of Large Language Models (LLMs) has catalyzed progress in multi-agent collaboration, enabling sophisticated distributed planning and communication strategies~\cite{bo2024reflective,guo2024embodied,nasiriany2024robocasa,zhou2023sotopia}. However, a primary limitation of these LLM-based approaches is their reliance on symbolic or textual representations, which are detached from the visual world. This detachment restricts their capacity to handle perceptual ambiguities and perform fine-grained spatial reasoning. While a few pioneering studies have started to integrate Vision--Language Models (VLMs) into multi-agent systems~\cite{wang2025rad,zhang2024combo,kang2025viki,kang2026advances,qinrobomonster}, they typically treat each agent's viewpoint in isolation or default to single-view reasoning. For instance, COMBO~\cite{zhang2024combo} assumes a shared global state rather than distributed ego-centric fusion, sidestepping the need for cross-view entity resolution. In stark contrast, our work directly confronts this limitation by proposing a framework centered on compositional, cross-view reasoning for deeper and more effective collaboration.

\paragraph{Spatial Understanding.}
Spatial understanding---the ability to parse intricate geometric configurations, spatial layouts, and object interrelations from diverse visual inputs---is critical for intelligent systems across a wide range of domains, from geometric problem-solving~\cite{gao2023g,shi2024math,zhang2024mathematical} to embodied robotics~\cite{hu2023look,ji2025robobrain}. To bolster the spatial reasoning capabilities of VLMs, recent works have explored increasingly advanced training methodologies. Techniques such as multi-stage supervision with Chain-of-Thought (CoT) prompting~\cite{xu2411llava,wei2022chain} and reinforcement learning (RL) with carefully engineered reward mechanisms~\cite{guo2025deepseek} have yielded significant gains. Subsequent research has further underscored the importance of highly tailored reward designs for complex visual reasoning tasks~\cite{liu2025noisyrollout,liu2025visual,tan2025reason}. MindCube~\cite{yin2025spatialmentalmodelinglimited} represents a notable step in this direction, showing that VLMs can infer hidden spatial layouts from a few static views. However, their setting fundamentally differs from ours. While it focuses on mental reconstruction of hidden layouts from a few static viewpoints, our work targets cross-view reasoning for embodied manipulation, where multiple distributed robots contribute heterogeneous ego-centric observations over time. We further study this capability across diverse embodied task settings, as detailed in Section~\ref{sec:task_overview}. Building upon this line of inquiry, our work addresses the distinct and even more complex challenge of \textit{multi-agent} spatial understanding. Here, the central problem is to synthesize a globally coherent and semantically consistent scene representation from fragmented, ego-centric observations. This necessitates not only unifying disparate multi-view visual data but also enforcing strong and robust spatial consistency across multiple, partially overlapping perspectives.

\paragraph{Reinforcement Learning for Visual Reasoning.}
Reinforcement Learning (RL) has emerged as a powerful and versatile paradigm for training intelligent agents that reason from high-dimensional visual data~\cite{zhang2025landscape}. Moving beyond the static nature of supervised pretraining, RL enables models to learn through direct and interactive environmental engagement, optimizing their policies via reward-driven feedback~\cite{liu2025noisyrollout,tan2025reason,sarch2025grounded,chen2025learning}. Foundational applications in perception-driven tasks, such as navigation~\cite{zhu2017target} and manipulation~\cite{zhou2025roborefer,kang2025viki}, have demonstrated RL's strong efficacy in tightly coupling perception and action. More recently, the field has progressed toward structured reward designs for multi-modal reasoning, incorporating mechanisms like CoT guidance~\cite{zhang2024improve} and geometric consistency constraints~\cite{jiang2024visual} to further enhance embodied decision-making performance. Despite these advances, the predominant focus of RL-based visual reasoning has remained on single-agent, single-image settings, leaving multi-view reasoning largely underexplored. Recent efforts such as VLA-R1~\cite{ye2025vlar1} and VLM-R1~\cite{shen2025vlmr1} apply RL to vision-language models, yet they operate exclusively on single-agent policies without cross-view reasoning or distributed observation fusion. Moreover, existing RL formulations rarely address how visual evidence from multiple ego-centric viewpoints should be aligned, fused, and jointly grounded. Our framework extends this paradigm to the multi-agent, multi-view context, introducing a novel and unified reward structure designed explicitly to foster collaborative spatial reasoning.
% \paragraph{Reinforcement Learning in Spatial Reasoning}
% Reinforcement learning (RL) has recently emerged as a powerful paradigm for enhancing spatial reasoning in embodied agents. Beyond supervised pretraining, RL enables models to explore and adapt to diverse environments through reward-driven optimization. Early works applied RL to perception-driven navigation and manipulation tasks~\cite{}, demonstrating its effectiveness in bridging perception with action. More recent advances incorporate structured reward designs to improve multi-modal reasoning: for instance, reinforcement learning with chain-of-thought guidance~\cite{} or geometric consistency constraints~\cite{} significantly boosts embodied decision-making and visual grounding. In VLM-based systems, RL has also been shown to refine object localization and enhance scene understanding by integrating spatial priors into the reward signal~\cite{}.

\section{Ego-to-World Task}
    \begin{figure}[t]
    \centering
    \includegraphics[width=1\linewidth]{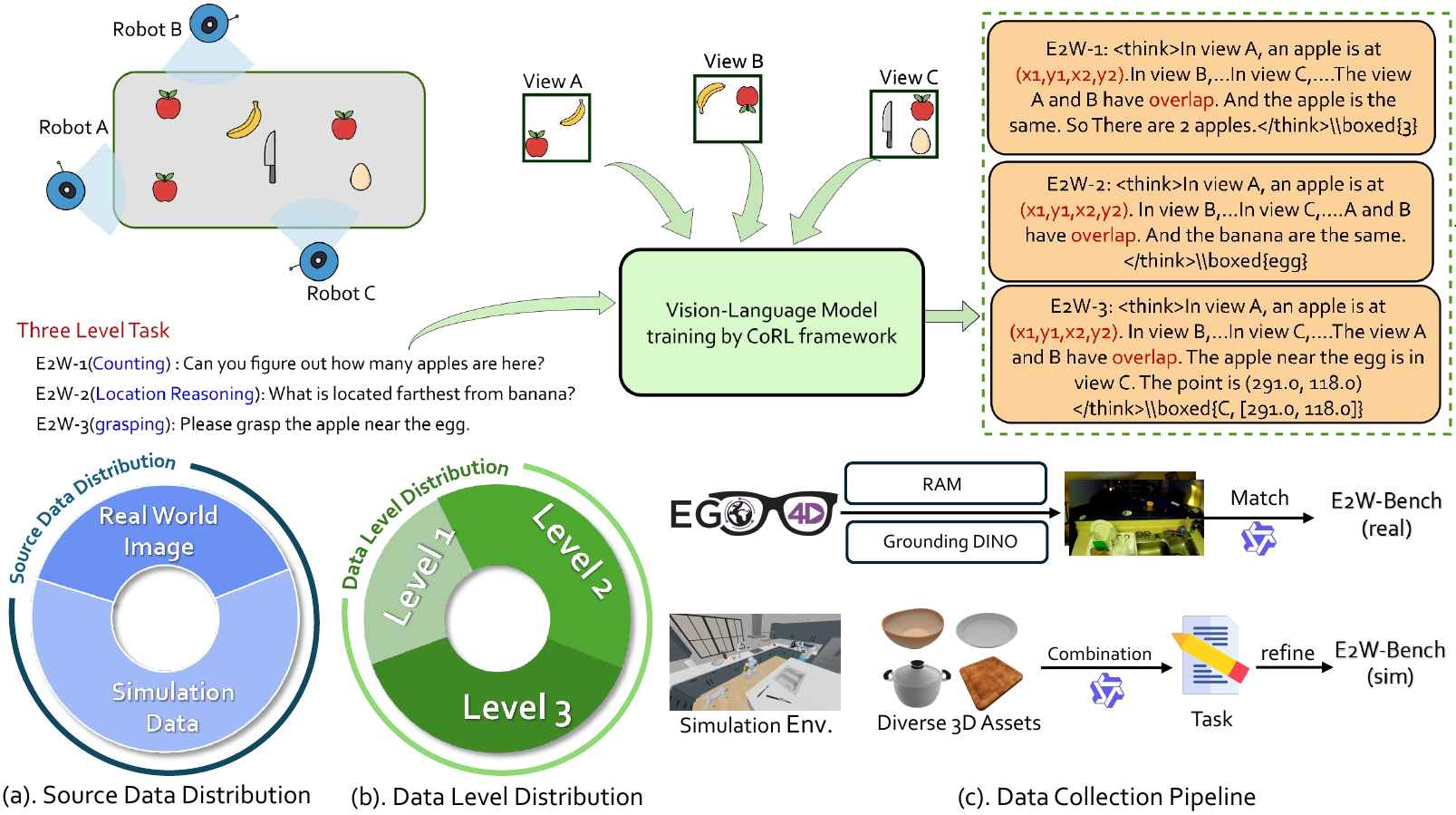}
    \caption{\textbf{Overview of the Ego-to-World (E2W) Benchmark.}
    \textbf{Top:} Multiple agents (Robot A, B, C) each provide partial ego-centric views of a shared scene.
    The vision language model trained with our CoRL framework integrate these complementary perspectives to solve three tasks:
    Counting (E2W-1), Location Reasoning (E2W-2), and Grasping (E2W-3).
    \textbf{Bottom:} The benchmark combines diverse real and simulated data and organizes them into varying complexity levels.}
    \label{fig:corl-data}
\end{figure}

\label{sec:task_overview}
To systematically evaluate collaborative spatial reasoning, we introduce the \emph{Ego-to-World Benchmark} (\textbf{E2W-Bench}), which operationalizes a multi-agent, multi-view paradigm. 
As shown in Figure~\ref{fig:corl-data}, multiple robotic agents capture partial ego-centric observations of a shared 3D environment. 
The central challenge for a Vision--Language Model is to integrate these fragmented perspectives into a coherent global scene representation and to answer natural-language queries or perform action-oriented predictions. 
E2W-Bench consists of two categories of tasks: spatial reasoning QA (E2W-1, E2W-2) and perception for grasping (E2W-3). Further dataset details are provided in Appendix~\ref{app:E2W}.

\textbf{E2W-1 (Counting).}  
This task evaluates the ability to aggregate object instances across overlapping views and output an accurate global count.

\textbf{E2W-2 (Location Reasoning).}  
Here the model must infer spatial relations among objects that never co-occur in a single view, requiring cross-view reasoning to answer correctly in natural language.

\textbf{E2W-3 (Grasping).}  
Unlike the QA tasks, E2W-3 requires action-oriented predictions. The model must translate a language command involving spatial relations into precise 2D coordinates within specific agent’s viewpoint, thereby linking compositional reasoning to downstream robotic manipulation.

% Together, these tasks provide a comprehensive benchmark that evaluates both high-level symbolic reasoning and embodied spatial referring, setting the stage for our CoRL framework.

Together, these tasks provide a comprehensive benchmark that jointly evaluates high-level symbolic reasoning, fine-grained visual grounding, and embodied spatial referring across diverse scenarios. E2W-Bench comprises over 100k simulated and 60k real-world samples spanning more than 15k scenes and 50+ object categories. Train and test sets are scene-disjoint, ensuring that no scene configuration appears in both splits. By explicitly linking abstract reasoning to actionable predictions, E2W-Bench offers a rigorous testbed for multi-view understanding.

\section{Methods}
    
\subsection{Overview}

\begin{figure}[t]
    \centering
    \includegraphics[width=1.0\linewidth]{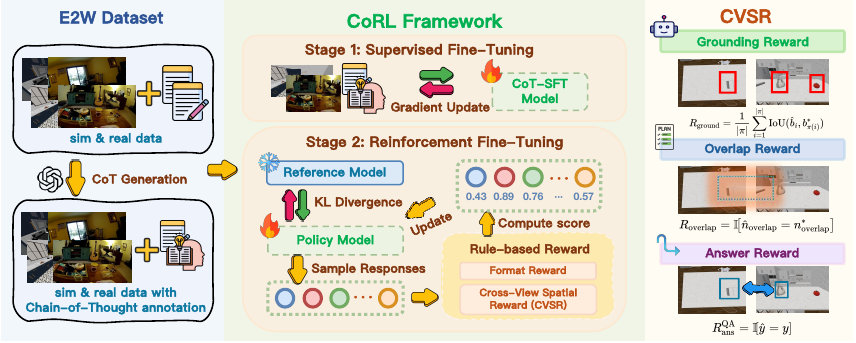}
    \caption{\textbf{CoRL framework.} The model is first initialized via supervised fine-tuning (SFT) on Chain-of-Thought annotations, then refined with reinforcement learning (RL). During RL, the policy is optimized with an format reward and the Cross-View Spatial Reward (CVSR), which supplies dense feedback on cross-view fusion and spatial consistency, guiding robust collaborative reasoning.}
    \label{fig:framework}
\end{figure}

We present \textbf{\textit{CoRL}} (Collaborative Spatial Reasoning Reinforcement Learning), a framework that equips Vision--Language Models (VLMs) with collaborative perception and reasoning capabilities in multi-agent embodied settings. A central VLM aggregates and reasons over partial, ego-centric observations streamed by multiple agents, producing task-specific outputs under a unified interface. The overall architecture is shown in Figure~\ref{fig:framework}.

\subsection{Problem Formulation}

Consider a shared environment $\mathcal{E}$ populated by $N$ embodied agents $\mathcal{A}=\{a_1,\dots,a_N\}$. At a given time, each agent $a_i$ observes an ego-centric RGB image $I_i\in\mathbb{R}^{H\times W\times 3}$. All agents receive a common natural-language query $Q$. A central VLM implements a policy $\pi_\theta$ parameterized by $\theta$ that consumes the multi-view input and the query to produce a prediction
\begin{equation}
    \hat{y}=\pi_\theta\!\big(\{I_i\}_{i=1}^{N},\,Q\big),
    \label{eq:policy}
\end{equation}
where $\hat{y}$ is either a textual response (for counting and relational reasoning) or a pair of image-plane coordinates (for grasping).

Training is cast as maximizing the expected task reward over a dataset $\mathcal{D}$ of instances $(\{I_i\},Q,y)$ with ground-truth label $y$:
\begin{equation}
    \max_{\theta}\;\mathbb{E}_{(\{I_i\},Q,y)\sim\mathcal{D}}\big[\,R(\hat{y},y)\,\big].
    \label{eq:objective}
\end{equation}
The reward function $R$ evaluates the quality of $\hat{y}$ against $y$ and is central to our method; its design is detailed in Section~\ref{sec:reward_design}.

\subsection{CoRL Training Pipeline}
\label{sec:training_pipeline}

CoRL is trained in two stages: supervised fine-tuning for cold start, followed by reinforcement learning.

\paragraph{Supervised Fine-Tuning.}
We initialize $\pi_\theta$ with SFT on CoT-augmented data. Each training tuple is $(x,q,r,a)$, where $x$ denotes the multi-view inputs, $q$ is the query, $r$ is the intermediate reasoning trace, and $a$ is the final answer. Let $y=[r,a]$ be the concatenated target sequence. The SFT objective is the usual next-token log-likelihood:
\begin{equation}
\mathcal{L}_{\mathrm{SFT}}
= -\,\mathbb{E}_{(x,q,r,a)\sim\mathcal{D}}
\sum_{t=1}^{|y|}\log \pi_{\theta}\!\big(y_t \mid x, q, y_{<t}\big).
\label{eq:sft}
\end{equation}
This yields an initial policy $\pi_{\mathrm{SFT}}$ (denoted $\pi_0$) that provides a strong starting point for RL.

\paragraph{Reinforcement Learning Fine-Tuning.}
After SFT initialization, we further optimize the policy with \emph{Group Relative Policy Optimization} (GRPO).
GRPO improves stability and sample efficiency by normalizing rewards within each sampled group of candidate responses, thereby computing \emph{group-relative advantages} that reduce variance and sharpen credit assignment.

Concretely, for an input $u=(x,q)$, we draw $G$ candidate responses $\{y_j\}_{j=1}^{G}$ from the current policy $\pi_\theta$ and score each with a reward $R_j$ (Section~\ref{sec:reward_design}).
GRPO computes the empirical mean $\bar{R}$ and standard deviation $\sigma_R$ of the rewards and defines a standardized advantage for each candidate:
\begin{equation}
\mathcal{A}_j = \frac{R_j - \bar{R}}{\sigma_R}.
\label{eq:adv_new}
\end{equation}
This group-relative normalization emphasizes responses that outperform their peers, rather than absolute reward magnitude, and thus stabilizes training.

To update the policy, we first define the probability ratio between the current policy $\pi_\theta$ and the reference policy $\pi_0$ from SFT:
\begin{equation}
r_j(\theta) = \frac{\pi_{\theta}(y_j\mid u)}{\pi_{0}(y_j\mid u)}.
\label{eq:prob_ratio}
\end{equation}
The core of the GRPO objective is a clipped surrogate function, which constrains the policy update step size. This objective is defined as:
\begin{equation}
\mathcal{L}^{\mathrm{CLIP}}(\theta) = \mathbb{E}_{u}\Bigg[\sum_{j=1}^{G} \min\Big(r_j(\theta)\mathcal{A}_j, \;\mathrm{clip}\big(r_j(\theta), 1-\epsilon, 1+\epsilon\big)\mathcal{A}_j\Big)\Bigg],
\label{eq:grpo_clip}
\end{equation}
where $\epsilon$ is a hyperparameter that defines the clipping range.

The final policy parameters are updated by maximizing the full GRPO objective, which incorporates a KL divergence penalty to further regulate the policy update:
\begin{equation}
\mathcal{J}(\theta)
 = \mathcal{L}^{\mathrm{CLIP}}(\theta)
   - \beta\,\mathrm{D_{KL}}\!\big(\pi_{\theta}(\cdot\mid u)\,\|\,\pi_{0}(\cdot\mid u)\big),
\label{eq:grpo_final}
\end{equation}
where $\pi_{0}$ is the SFT-initialized reference policy and $\beta>0$ regulates the trust region enforced by the KL regularizer.
This formulation allows CoRL to directly optimize spatial reasoning rewards while preserving stability and sample efficiency, fully leveraging the group-relative advantage mechanism within a robust PPO-style optimization framework.

\subsection{Cross-View Spatial Reward (CVSR) Design}
\label{sec:reward_design}

The total reward combines an output-format component with a cross-view spatial component:
\begin{equation}
R=\lambda_{1}\,R_{\mathrm{format}}+\lambda_{2}\,R_{\mathrm{CVSR}},
\qquad \lambda_{1},\lambda_{2}>0.
\label{eq:total_reward}
\end{equation}

\paragraph{Output-Format Reward $R_{\mathrm{format}}$.}
To ensure interpretability and reliable parsing, the model receives a binary reward for structural correctness. Specifically, the intermediate reasoning must be enclosed in \texttt{<think>...</think>} tags and the final answer must appear in a designated box; success yields $R_{\mathrm{format}}=1$, otherwise $0$. This encourages the model to articulate a reasoning trace prior to committing to an answer.

\paragraph{Cross-View Spatial Reward $R_{\mathrm{CVSR}}$.}
CVSR delivers dense feedback targeted at collaborative spatial reasoning. It aggregates three components:
\begin{equation}
R_{\mathrm{CVSR}}
= w_{\mathrm{ground}}\,R_{\mathrm{ground}}
+ w_{\mathrm{overlap}}\,R_{\mathrm{overlap}}
+ w_{\mathrm{ans}}\,R_{\mathrm{ans}},
\label{eq:cvsr_decomp}
\end{equation}
with nonnegative weights $w_{\mathrm{ground}}, w_{\mathrm{overlap}}, w_{\mathrm{ans}}$.

\emph{(i) Grounding reward $R_{\mathrm{ground}}$.}
To align reasoning with visual evidence, the model is prompted to emit bounding boxes for key objects referenced in its chain of thought. Let $\hat{B}=\{\hat{b}_i\}_{i=1}^{m}$ be predicted boxes and $B^\ast=\{b^\ast_j\}_{j=1}^{n}$ ground-truth boxes. We compute an optimal bipartite matching $\sigma$ via the Hungarian algorithm that maximizes total IoU, and define
\begin{equation}
R_{\mathrm{ground}}
= \frac{1}{|\sigma|}\sum_{i=1}^{|\sigma|}\mathrm{IoU}\big(\hat{b}_i,\,b^\ast_{\sigma(i)}\big),
\end{equation}
which provides a dense localization signal.

\emph{(ii) Overlap accuracy $R_{\mathrm{overlap}}$.}
To incentivize cross-view entity resolution, the model must report the number of unique object instances that appear in more than one view, denoted $\hat{n}_{\mathrm{overlap}}$. Comparing to the ground truth $n^\ast_{\mathrm{overlap}}$ yields
\begin{equation}
R_{\mathrm{overlap}}
=\mathbb{I}\!\left[\hat{n}_{\mathrm{overlap}}=n^\ast_{\mathrm{overlap}}\right],
\end{equation}
encouraging the model to distinguish redundant from complementary observations before global aggregation.

\emph{(iii) Answer correctness $R_{\mathrm{ans}}$.}
This term evaluates task completion and is defined per task type. For counting and location reasoning (textual outputs),
\begin{equation}
R_{\mathrm{ans}}^{\mathrm{QA}}
=\mathbb{I}\!\left[\hat{y}=y\right].
\end{equation}
For grasping (coordinate output $\hat{y}=(\hat{u},\hat{v})$ with ground truth $y=(u,v)$), we use a distance-shaped reward
\begin{equation}
R_{\mathrm{ans}}^{\mathrm{grasp}}
=\max\!\Big(0,\;1-\tfrac{\lVert \hat{y}-y\rVert_2}{d_{\max}}\Big),
\end{equation}
where $d_{\max}$ is a normalization radius. While $R_{\mathrm{ground}}$ and $R_{\mathrm{overlap}}$ shape intermediate spatial reasoning, $R_{\mathrm{ans}}$ enforces correctness of the final output.

\section{Experiments}
    \label{sec:experiments}
\subsection{Experimental Setup}

\paragraph{Simulation Environment and Data Generation.}
The simulated portion of E2W-Bench is constructed using RoboFactory~\cite{qin2025robofactory}, a framework for embodied multi-agent manipulation built on ManiSkill3~\cite{tao2024maniskill3}. Chain-of-Thought (CoT) annotations for supervised fine-tuning are generated through a three-step pipeline: ground-truth bounding boxes are first extracted from simulator metadata, spatial relations are then computed programmatically, and finally reasoning traces are assembled from structured templates. This ensures high-quality, consistent annotations without relying on manual labeling.

\paragraph{Baselines and Protocol.}
We compare \textbf{CoRL} with (i) proprietary VLMs: GPT-5~\cite{openai2025gpt5}, Gemini-2.5-Pro~\cite{comanici2025gemini}, and Doubao-Seed-1.6~\cite{bytedance_seed1_6}; (ii) open-source VLMs: GLM-4.5V~\cite{vteam2025glm45vglm41vthinkingversatilemultimodal}, Qwen2.5-VL-32B~\cite{bai2025qwen2}, LLaMA-3.2-11b-vision-instruct~\cite{dubey2024llama}, SpaceQwen3B~\cite{yang2025qwen3}; and (iii) fine-tuned variants on Qwen2.5-VL-Instruct~\cite{bai2025qwen2} backbones (3B/7B): +SFT, +RL-ZERO (RL from scratch). All baseline models are evaluated with zero-shot inference under identical prompts and input aggregation protocol, ensuring a fair comparison. Complete training details and prompt templates are provided in the Appendix.

\paragraph{Implementation Details.}
All models are built on the Qwen2.5-VL-Instruct backbone. During the SFT stage, we train for 3 epochs with a cosine-decay learning rate schedule (peak $2\!\times\!10^{-5}$) and a per-device batch size of 4, accumulating gradients over 4 steps. For RL fine-tuning with GRPO, we sample $G\!=\!8$ candidate responses per input, set the clipping range $\epsilon\!=\!0.2$, the KL coefficient $\beta\!=\!0.04$, and the reward weights $\lambda_1\!=\!0.1$, $\lambda_2\!=\!1.0$. The normalization radius $d_{\max}$ for the grasping reward is set to $100$ pixels. RL training runs for 200 steps with a learning rate of $1\!\times\!10^{-6}$. All experiments are conducted on 8$\times$NVIDIA H200.

\paragraph{Evaluation Benchmarks and Metrics.}
We conduct our primary evaluation on the proposed \textbf{E2W-Bench}, which comprises the Counting, Location Reasoning, and Grasping tasks detailed in Section~\ref{sec:task_overview}. For the QA-based tasks (Counting and Location Reasoning), we report exact match accuracy. For Grasping, we report a normalized score from 0 to 100, calculated based on the Euclidean distance between the predicted and ground-truth coordinates, consistent with the task's reward function defined in our methodology. To further assess the generalization capabilities of our approach, we also report performance on the external Where2Place dataset, a standard testbed for spatial reasoning. A detailed description of dataset statistics, implementation specifics, and evaluation protocols is provided in the Appendix~\ref{app:E2W}.

\subsection{Main Results on E2W-Bench}

\def\best#1{\textbf{#1}}
\def\second#1{\underline{#1}}

\begin{table}[t]
\centering
\caption{
\textbf{Performance on E2W-Bench.}
Tasks are grouped into Reasoning (accuracy, \%) and Perception (score, \%).
(S)~=~simulation, (R)~=~real-world.
\best{Best} in bold, \second{second best} underlined.
}
\label{tab:main_performance_final}
\resizebox{\linewidth}{!}{%
\setlength{\tabcolsep}{4pt}
\renewcommand{\arraystretch}{1.1}
\begin{tabular}{
  l
  S[table-format=2.1] S[table-format=2.1] S[table-format=2.1]
  S[table-format=2.2]
  S[table-format=2.2] S[table-format=2.2] S[table-format=2.2]
}
\toprule
\multirow{2}{*}{\textbf{Model}}
  & \multicolumn{4}{c}{\textbf{Reasoning (Acc)}}
  & \multicolumn{3}{c}{\textbf{Perception (Score)}} \\
\cmidrule(lr){2-5}
\cmidrule(lr){6-8}
& {\textbf{E2W-1}} & {\textbf{E2W-2(S)}} & {\textbf{E2W-2(R)}} & {\textbf{Avg.}}
& {\textbf{E2W-3(S)}} & {\textbf{E2W-3(R)}} & {\textbf{Avg.}} \\
\midrule
\multicolumn{8}{c}{\textit{Closed Source Models}} \\
\midrule
GPT-5           & 42.5 & 48.5 & 72.5 & 54.50 & 50.43 & 12.02 & 31.23 \\
Doubao-Seed-1.6 & 35.0 & 40.0 & 46.0 & 40.33 & 16.60 & 5.26  & 10.93 \\
Gemini-2.5-Pro  & 32.5 & 42.5 & 50.0 & 41.67 & 35.98 & 10.15 & 23.07 \\
\midrule
\multicolumn{8}{c}{\textit{Open Source Models}} \\
\midrule
GLM-4.5v        & 34.5 & 29.0 & 56.0 & 39.83 & 2.78  & 0.84  & 1.81 \\
SpaceQwen-3B    & 21.5 & 6.0  & 60.0 & 29.17 & 16.15 & 5.07  & 10.61 \\
LLaMA-3.2-11b  & 16.5 & 12.5 & 17.5 & 15.50 & 7.78 & 3.38 & 5.58 \\
Qwen2.5VL-32B   & 21.5 & 28.0 & 37.0 & 28.83 & 31.25 & 9.16 & 20.21 \\
\midrule
\multicolumn{8}{c}{\textit{Qwen2.5VL-3B}} \\
\midrule
Zero-shot    & 22.0 & 15.5 & 58.0 & 31.83 & 24.08 & 7.48 & 15.78 \\
+SFT          & 47.0 & 63.0 & 84.0 & 64.67 & 93.00 & \second{42.06} & 67.53 \\
+RL-ZERO      & 23.0 & 39.5 & 83.5 & 48.67 & 50.63 & 8.02  & 29.33 \\
+CoRL (Ours)  & \second{59.0} & 75.5 & \second{86.0} & \second{73.50} & \best{96.30} & 41.82 & \second{69.06} \\
\midrule
\multicolumn{8}{c}{\textit{Qwen2.5VL-7B}} \\
\midrule
Zero-shot     & 17.0 & 17.0 & 64.5 & 32.83 & 28.83 & 5.78 & 17.31 \\
+SFT          & 44.5 & \second{88.0} & 84.5 & 72.33 & 90.99 & 40.76 & 65.88 \\
+RL-ZERO     & 16.0 & 56.0 & 82.5 & 51.50 & 92.60 & 11.65 & 52.13 \\
+CoRL (Ours)  & \best{61.0} & \best{97.0} & \best{90.0} & \best{82.67} & \second{95.69} & \best{44.32} & \best{70.01} \\
\bottomrule
\end{tabular}
}
\end{table}

Our main findings on the E2W-Bench are summarized in Table~\ref{tab:main_performance_final}. The results show a clear performance hierarchy across the different model categories. In a zero-shot setting, proprietary models like GPT-5 establish the strongest baseline, significantly outperforming open-source VLMs, which generally struggle with the benchmark's complex multi-view demands. The model fine-tuned on E2W-Bench training set, demonstrate the efficacy of our proposed training pipeline. The SFT-only models establish a very strong performance level, while the full CoRL (SFT+RL) framework consistently achieves the highest scores across both 3B and 7B model scales. Notably, the RL-ZERO model performs poorly, underscoring the necessity of a supervised warm-up phase to stabilize optimization and provide a reliable initialization. This observation further highlights the complementary nature of SFT and RL, where supervised fine-tuning imparts essential reasoning priors and reinforcement learning subsequently refines them toward task-specific objectives.

\looseness=-1
A deeper analysis across task categories reveals distinct performance patterns. On the \textbf{Reasoning} tasks (E2W-1 and E2W-2), which require aggregating abstract information across views, the largest models show some inherent capability. However, the RL stage of our CoRL framework provides a crucial advantage, refining the model's ability to resolve cross-view ambiguities and synthesize a coherent global state, as evidenced by the superior scores of CoRL models. The most pronounced gap emerges in the \textbf{Perception Grounding} task (E2W-3). Here, nearly all zero-shot models fail to produce reliable coordinates, exposing a weakness in fine-grained spatial grounding. In contrast, our models excel, thanks to explicit supervision in our dataset and, for CoRL, the targeted policy optimization from CVSR's grounding and consistency rewards, which shape the model toward physically precise and reliable outputs.

\looseness=-1
We also note an important observation regarding the performance gap between simulated and real-world subsets. Scores on the real-world subset (R) are generally higher than on the simulated subset (S), which may appear counter-intuitive. This is by design: the simulation subset features deliberately challenging configurations including cluttered scenes, severe occlusions, and adversarial spatial layouts, whereas the real-world subset is curated with cleaner visibility. Thus the difficulty gap, rather than domain mismatch, primarily explains the observed score difference.

Beyond task-level accuracy, a critical capability that CoRL acquires is \emph{cross-view entity resolution}, determining when the same physical object appears across multiple views. Without this skill, models frequently double-count objects or fail to establish spatial relations that span views. The CVSR's overlap reward ($R_{\mathrm{overlap}}$) directly addresses this by incentivizing the model to track unique instances across views, learning to use shared visible objects as geometric anchors that link partial observations into a coherent global representation.

\subsection{Ablation Studies}

To validate our design choices, we conduct a series of ablation studies. We first analyze the contribution of each component within our CVSR design, and then investigate the performance gap between reasoning from distributed ego-centric views versus a single, privileged global view. These studies provide a deeper understanding of how individual reward signals and viewpoint configurations jointly influence the emergence of robust cross-view reasoning.

\begin{table}[t]
\centering
\begin{minipage}[t]{0.48\linewidth}
    \centering
    \small
    \setlength{\tabcolsep}{3pt}
    \renewcommand{\arraystretch}{1.2}
    \caption{\textbf{Ablation of CVSR Components}. Results of CoRL-7B on E2W-Bench.}
    \label{tab:reward_ablation}
    \begin{tabular}{lccc}
        \toprule
        \textbf{Setting} & \textbf{E2W-1} & \textbf{E2W-2(S)} & \textbf{E2W-3(S)} \\
        \midrule
        CoRL (CVSR)                 & \textbf{61.0} & \textbf{97.0} & \textbf{95.69} \\
        \midrule
        \multicolumn{4}{l}{\textit{Ablation}} \\
        \quad w/o $R_{\text{ans}}$     & 10.5 & 15.5 & 40.32 \\
        \quad w/o $R_{\text{ground}}$  & 50.5 & 90.5 & 74.32 \\
        \quad w/o $R_{\text{overlap}}$ & 56.5 & 90.0 & 84.31 \\
        \quad w/o $R_{\text{format}}$  & 58.5 & 93.0 & 89.31 \\
        \bottomrule
    \end{tabular}
\end{minipage}%
\hfill
\begin{minipage}[t]{0.48\linewidth}
    \centering
    \small
    \setlength{\tabcolsep}{6pt}
    \renewcommand{\arraystretch}{1.2}
    \caption{\textbf{Single-view vs.\ Multi-view}. Performance on E2W-Bench.}
    \label{tab:single_multi_view}
    \begin{tabular}{lcc}
        \toprule
        \textbf{Setting} & \textbf{E2W-1} & \textbf{E2W-2(S)} \\
        \midrule
        \multicolumn{3}{l}{\textit{7B Backbone}} \\
        \quad Single-view   & 34.0 & 54.0 \\
        \quad Multi-view    & \textbf{61.0} & \textbf{97.0} \\
        \midrule
        \multicolumn{3}{l}{\textit{3B Backbone}} \\
        \quad Single-view   & 36.5 & 51.5 \\
        \quad Multi-view    & \textbf{59.0} & \textbf{75.5} \\
        \bottomrule
    \end{tabular}
\end{minipage}
\end{table}

\paragraph{Impact of CVSR Components.}

We ablate each component of the Cross-View Spatial Reward (CVSR) to assess its individual contribution (Table~\ref{tab:reward_ablation}).
Removing the answer correctness reward ($R_{\text{ans}}$) causes a catastrophic drop in E2W-1 accuracy from \textbf{61.0\%} to 10.5\%, showing that intermediate shaping signals alone cannot ensure correct final solutions.
Eliminating the grounding reward ($R_{\text{ground}}$) reduces E2W-3(S) by over 21 points, confirming its role in aligning symbolic reasoning with precise object locations.
Dropping the overlap reward ($R_{\text{overlap}}$) weakens counting and relational reasoning by impairing cross-view object consistency.
Finally, the format reward ($R_{\text{format}}$) provides smaller but meaningful gains by enforcing structured outputs and stabilizing optimization.
\looseness=-1
Together these results show that only the full CVSR, balancing correctness, grounding, consistency, and structural integrity, can robustly guide multi-view reasoning and overall embodied performance.

\paragraph{Multi-Views vs.\ Single View Reasoning.}
\looseness=-1
We next investigate a fundamental question: is it more effective to reason from a single global view or from multiple ego-centric views captured by distributed agents? The single-view baseline uses a global overhead camera covering the entire workspace, simulating a typical single-robot setup with full scene visibility. Table~\ref{tab:single_multi_view} shows that the multi-view configuration consistently outperforms the single view despite the latter's complete coverage. For instance, on E2W-2(S), our multi-view approach achieves \textbf{97.0\%} accuracy, significantly surpassing the single view's 54.0\%. This gap arises because a single global viewpoint suffers from perspective distortion and lacks the fine-grained geometric detail that close-range ego-centric views provide. In contrast, multiple ego-centric views, captured from diverse positions, naturally provide complementary information and broader coverage of the scene, reducing ambiguity and enabling more complete spatial reasoning.

\subsection{External Benchmark: Where2Place}

\begin{table}[t]
\centering
\small
\setlength{\tabcolsep}{4pt}
\caption{\textbf{Left:} Where2Place results. \textbf{Right:} Real-world robot evaluation.}
\label{tab:where2place}
\begin{minipage}[c]{0.34\linewidth}
    \centering
    \footnotesize
    \renewcommand{\arraystretch}{1.42}
    \begin{tabular}{lc}
    \toprule
    \textbf{Model} & \textbf{Score}~$\uparrow$ \\
    \midrule
    SpaceLLaVA~\cite{foutter2024space} & 11.8 \\
    RoboPoint~\cite{yuan2024robopoint}  & 46.8 \\
    Molmo-7B~\cite{deitke2025molmo}   & 45.0 \\
    \midrule
    \textbf{CoRL-7B} & \textbf{50.9} \\
    \bottomrule
    \end{tabular}
\end{minipage}%
\hfill
\begin{minipage}[c]{0.63\linewidth}
    \centering
    \footnotesize
    \renewcommand{\arraystretch}{1.15}
    \setlength{\tabcolsep}{3pt}%
    \begin{tabular}{m{3.6cm}|cc}
    \toprule
    \textbf{Task} & \textbf{RoboPoint} & \textbf{CoRL-7B} \\
    \midrule
  Grasp blue block farthest from strawberry, place on red block & 0.0 & \textbf{65.0} \\

    Pick carambola aligned with banana, place on red block & 0.0 & \textbf{30.0} \\
    \bottomrule
    \end{tabular}%
\end{minipage}
\end{table}

We further test single-image spatial grounding on \textbf{Where2Place}~\cite{yuan2024robopoint}, which evaluates fine-grained point prediction from a single RGB view (Table~\ref{tab:where2place}, left).
CoRL-7B attains 50.9 points, outperforming RoboPoint~\cite{yuan2024robopoint} (46.8) and Molmo-7B~\cite{deitke2025molmo} (45.0).
This shows that cross-view training with CVSR not only benefits multi-view reasoning but also preserves strong single-image pointing ability.

\subsection{Real-World Evaluation}

\begin{figure}[t]
    \centering
    \includegraphics[width=\linewidth]{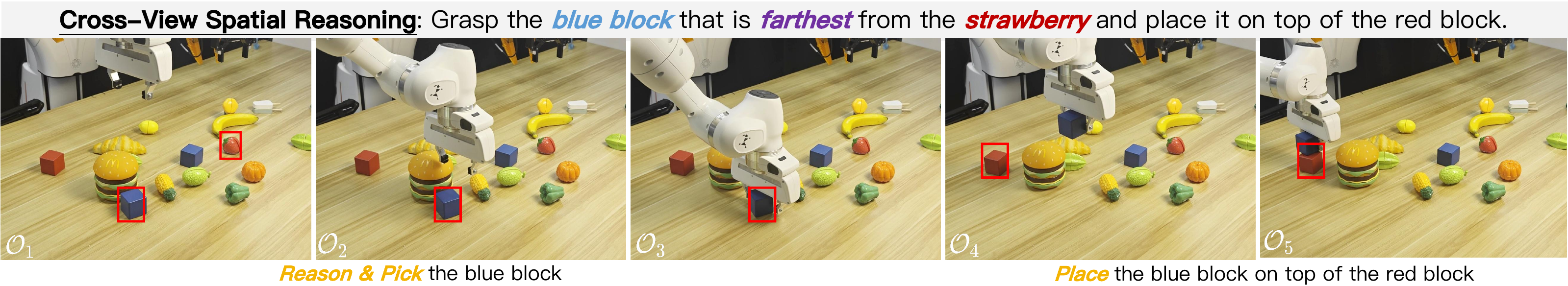}
    \caption{Illustrative demonstrations of our model in real-world robotic manipulation tasks involving pick-and-place with cross-view spatial reasoning.}
    \label{fig:real-world}
\end{figure}

We validate the effectiveness of CoRL-7B through real-world experiments. The platform comprises two Franka Research~3 arms and one Realman mobile base, each equipped with an Intel RealSense D435 RGB-D camera with calibrated extrinsics. All cameras are pre-calibrated to a shared world frame; during inference, CoRL-7B receives only the RGB images and the natural-language instruction, and predicts 2D target coordinates in a designated agent's image. These coordinates are then lifted to 3D workspace positions using the camera's depth channel and known intrinsics/extrinsics for downstream grasp execution. Importantly, the model itself never accesses depth or calibration data, it reasons purely from multi-view RGB inputs.

As shown in Table~\ref{tab:where2place} (right), our model achieves success rates of 65\% on the blue block picking task and 30\% on the carambola alignment task, significantly outperforming the RoboPoint~\cite{yuan2024robopoint} baseline (0\% on both tasks). These results highlight the robustness of our method in handling visual ambiguity, such as occlusions and distractors with similar colors. In particular, CoRL maintains stable predictions under viewpoint shifts, depth noise, and partial occlusions, reflecting its ability to integrate heterogeneous multi-agent observations and preserve spatial consistency in real deployments.

A key challenge in the real-world setting is bridging the sim-to-real gap. Real-world scenes introduce lighting variations, reflective surfaces, and sensor noise absent from simulation. We observe that the primary failure mode is not incorrect spatial reasoning but imprecise coordinate prediction: the model correctly identifies the target object and its approximate region, yet the predicted point occasionally falls on the object boundary. This suggests that tighter grounding supervision and resolution-aware coordinate regression are promising directions.
Qualitative rollouts in Figure~\ref{fig:real-world} further illustrate successful cross-view reasoning. Notably, in the blue block task, the target is partially occluded in one agent's view but fully visible from another; CoRL leverages both perspectives to resolve the ambiguity and produce a consistent prediction.

\subsection{Discussion}
Our comprehensive experiments convey a clear message: while model scale and pretraining provide a foundation, they are insufficient for mastering the complexities of collaborative embodied reasoning. The success of our SFT$\rightarrow$RL pipeline highlights a crucial insight: robust spatial intelligence emerges from a structured curriculum that first bootstraps foundational, in-domain knowledge via SFT, and then refines nuanced collaborative behaviors through targeted reward shaping via CVSR. The reinforcement learning stage proves essential for teaching the model not just \emph{what} the right answer is, but \emph{how} to systematically derive it by grounding its reasoning in visual evidence and maintaining cross-view consistency. This is further evidenced by the RL-ZERO baseline, which attempts RL from scratch and performs poorly, demonstrating that without the structured initialization from SFT, the reward landscape is too sparse for effective policy learning. Furthermore, the ablation on input modality (Table~\ref{tab:single_multi_view}) reveals a non-obvious trade-off between information density and contextual breadth. Under fixed computational budgets, multiple high-resolution local views prove more valuable than a single coarse global one, because close-range ego-centric views provide rich, complementary geometric detail that a single overhead camera, despite full workspace coverage, cannot capture due to perspective distortion. This suggests that future multi-robot systems should prioritize distributed sensing over monolithic viewpoints.

% \section{Ablation Studies}
%     \input{sections/ablation}

\section{Conclusion}
    \label{sec:conclusion}
We introduced \textbf{CoRL}, a two-stage SFT$\rightarrow$RL framework that equips Vision Language Models with collaborative spatial reasoning abilities from distributed ego-centric views. At the core is the Cross-View Spatial Reward (CVSR), which provides structured signals for multi-view fusion, cross-view consistency, and fine-grained grounding. Together with \textbf{E2W-Bench}, our work offers a unified testbed and methodology for studying collaborative perception in multi-agent embodied settings.

\looseness=-1
Extensive experiments demonstrate that CoRL consistently outperforms strong proprietary and open-source baselines. Ablation studies confirm that each CVSR component plays a complementary role, and that distributed ego-centric views consistently surpass a single global overhead view, validating that multi-agent perception, when properly fused through compositional reasoning, yields richer spatial understanding than any single privileged viewpoint. CoRL further generalizes to Where2Place and to real-world robotic manipulation. As embodied AI moves toward multi-robot deployments, we hope E2W-Bench and CoRL serve as a foundation for future research.

\section*{Ethics Statement}
This work makes use of the publicly available Ego4D~\cite{grauman2022ego4d} dataset and the ManiSkill3 simulation platform~\cite{tao2024maniskill3}. The dataset was collected and released in accordance with ethical research practices, and all experiments conducted in simulation do not involve human or animal subjects. Our study focuses on improving the capabilities of vision-language models, and we do not foresee any direct ethical concerns, including issues of privacy, safety, or fairness. We believe the potential benefits of advancing embodied AI research outweigh the minimal risks, and no conflicts of interest or ethical violations are associated with this work.

\section*{Reproducibility Statement}
We have taken extensive measures to ensure the reproducibility of our work. All code related to data preprocessing, model training, and evaluation is released at \href{https://github.com/hengzzzhou/CoRL_code}{\textsc{CoRL}}, and the trained models will be hosted on Hugging Face for public access. The processed data used in our experiments will also be made available to facilitate replication. Furthermore, detailed experimental settings, including hyperparameters, training schedules, and environment configurations, are documented in the appendix and supplementary materials.

\bibliography{iclr2026_conference}

@article{zhang2025landscape,
  title={The landscape of agentic reinforcement learning for llms: A survey},
  author={Zhang, Guibin and Geng, Hejia and Yu, Xiaohang and Yin, Zhenfei and Zhang, Zaibin and Tan, Zelin and Zhou, Heng and Li, Zhongzhi and Xue, Xiangyuan and Li, Yijiang and others},
  journal={TMLR},
  year={2025}
}

@article{kang2026advances,
  title={Advances and Innovations in the Multi-Agent Robotic System (MARS) Challenge},
  author={Kang, Li and Zhou, Heng and Song, Xiufeng and Li, Rui and Chen, Bruno NY and Wang, Ziye and Meng, Ximeng and Tao, Stone and Qin, Yiran and Liu, Xiaohong and others},
  journal={arXiv preprint arXiv:2601.18733},
  year={2026}
}

@article{qinrobomonster,
  title={RoboMonster: Compositional Generalization of Heterogeneous Multi-End Effector Embodied Agents},
  author={Qin, Yiran and Zhang, Zhemeng and Zhou, Heng and Kang, Li and Chen, Bruno NY and Meng, Ximeng and Song, Xiufeng and Ma, Jiahua and Yin, Zhenfei and Liu, Xiaohong and others},
  journal={arXiv preprint},
  year={2025}
}

@article{qinbuilding,
  title={Building Scalable Real-World Robot Data Generation via Compositional Simulation},
  author={Qin, Yiran and Ma, Jiahua and Kang, Li and Li, Wenzhan and Song, Xiufeng and Zhou, Heng and Yu, Jiwen and Yin, Zhenfei and Liu, Xihui and Torr, Philip and others},
  journal={arXiv preprint},
  year={2025}
}

@article{fang2023anygrasp,
  title={Anygrasp: Robust and efficient grasp perception in spatial and temporal domains},
  author={Fang, Hao-Shu and Wang, Chenxi and Fang, Hongjie and Gou, Minghao and Liu, Jirong and Yan, Hengxu and Liu, Wenhai and Xie, Yichen and Lu, Cewu},
  journal={IEEE Transactions on Robotics},
  volume={39},
  number={5},
  pages={3929--3945},
  year={2023},
  publisher={IEEE}
}

@article{yang2025qwen3,
  title={Qwen3 technical report},
  author={Yang, An and Li, Anfeng and Yang, Baosong and Zhang, Beichen and Hui, Binyuan and Zheng, Bo and Yu, Bowen and Gao, Chang and Huang, Chengen and Lv, Chenxu and others},
  journal={arXiv preprint arXiv:2505.09388},
  year={2025}
}

@article{dubey2024llama,
  title={The llama 3 herd of models},
  author={Dubey, Abhimanyu and Jauhri, Abhinav and Pandey, Abhinav and Kadian, Abhishek and Al-Dahle, Ahmad and Letman, Aiesha and Mathur, Akhil and Schelten, Alan and Yang, Amy and Fan, Angela and others},
  journal={arXiv e-prints},
  pages={arXiv--2407},
  year={2024}
}

@misc{vteam2025glm45vglm41vthinkingversatilemultimodal,
      title={GLM-4.5V and GLM-4.1V-Thinking: Towards Versatile Multimodal Reasoning with Scalable Reinforcement Learning}, 
      author={V Team and Wenyi Hong and Wenmeng Yu and Xiaotao Gu and Guo Wang and Guobing Gan and Haomiao Tang and Jiale Cheng and Ji Qi and Junhui Ji and Lihang Pan and Shuaiqi Duan and Weihan Wang and Yan Wang and Yean Cheng and Zehai He and Zhe Su and Zhen Yang and Ziyang Pan and Aohan Zeng and Baoxu Wang and Bin Chen and Boyan Shi and Changyu Pang and Chenhui Zhang and Da Yin and Fan Yang and Guoqing Chen and Jiazheng Xu and Jiale Zhu and Jiali Chen and Jing Chen and Jinhao Chen and Jinghao Lin and Jinjiang Wang and Junjie Chen and Leqi Lei and Letian Gong and Leyi Pan and Mingdao Liu and Mingde Xu and Mingzhi Zhang and Qinkai Zheng and Sheng Yang and Shi Zhong and Shiyu Huang and Shuyuan Zhao and Siyan Xue and Shangqin Tu and Shengbiao Meng and Tianshu Zhang and Tianwei Luo and Tianxiang Hao and Tianyu Tong and Wenkai Li and Wei Jia and Xiao Liu and Xiaohan Zhang and Xin Lyu and Xinyue Fan and Xuancheng Huang and Yanling Wang and Yadong Xue and Yanfeng Wang and Yanzi Wang and Yifan An and Yifan Du and Yiming Shi and Yiheng Huang and Yilin Niu and Yuan Wang and Yuanchang Yue and Yuchen Li and Yutao Zhang and Yuting Wang and Yu Wang and Yuxuan Zhang and Zhao Xue and Zhenyu Hou and Zhengxiao Du and Zihan Wang and Peng Zhang and Debing Liu and Bin Xu and Juanzi Li and Minlie Huang and Yuxiao Dong and Jie Tang},
      year={2025},
      eprint={2507.01006},
      archivePrefix={arXiv},
      primaryClass={cs.CV},
}

@misc{openai2025gpt5,
  author       = {OpenAI},
  title        = {Introducing GPT-5},
  year         = {2025},
  month        = {August},
  day          = {7},
}

@online{bytedance_seed1_6,
  author       = {ByteDance Seed},
  title        = {Seed 1.6},
  year         = {2025},
  url          = {https://seed.bytedance.com/en/seed1_6},
}

@article{comanici2025gemini,
  title={Gemini 2.5: Pushing the frontier with advanced reasoning, multimodality, long context, and next generation agentic capabilities},
  author={Comanici, Gheorghe and Bieber, Eric and Schaekermann, Mike and Pasupat, Ice and Sachdeva, Noveen and Dhillon, Inderjit and Blistein, Marcel and Ram, Ori and Zhang, Dan and Rosen, Evan and others},
  journal={arXiv preprint arXiv:2507.06261},
  year={2025}
}

@inproceedings{caron2021emerging,
  title={Emerging properties in self-supervised vision transformers},
  author={Caron, Mathilde and Touvron, Hugo and Misra, Ishan and J{\'e}gou, Herv{\'e} and Mairal, Julien and Bojanowski, Piotr and Joulin, Armand},
  booktitle={Proceedings of the IEEE/CVF international conference on computer vision},
  pages={9650--9660},
  year={2021}
}

@article{zhang2023recognize,
  title={Recognize Anything: A Strong Image Tagging Model},
  author={Zhang, Youcai and Huang, Xinyu and Ma, Jinyu and Li, Zhaoyang and Luo, Zhaochuan and Xie, Yanchun and Qin, Yuzhuo and Luo, Tong and Li, Yaqian and Liu, Shilong and others},
  journal={arXiv preprint arXiv:2306.03514},
  year={2023}
}

@article{zhu2022viola,
  title={VIOLA: Imitation Learning for Vision-Based Manipulation with Object Proposal Priors},
  author={Zhu, Yifeng and Joshi, Abhishek and Stone, Peter and Zhu, Yuke},
  journal={arXiv preprint arXiv:2210.11339},
  doi={10.48550/arXiv.2210.11339},
  year={2022}
}

@article{ravi2024sam,
  title={Sam 2: Segment anything in images and videos},
  author={Ravi, Nikhila and Gabeur, Valentin and Hu, Yuan-Ting and Hu, Ronghang and Ryali, Chaitanya and Ma, Tengyu and Khedr, Haitham and R{\"a}dle, Roman and Rolland, Chloe and Gustafson, Laura and others},
  journal={arXiv preprint arXiv:2408.00714},
  year={2024}
}

@article{kang2025viki,
  title={Viki-r: Coordinating embodied multi-agent cooperation via reinforcement learning},
  author={Kang, Li and Song, Xiufeng and Zhou, Heng and Qin, Yiran and Yang, Jie and Liu, Xiaohong and Torr, Philip and Bai, Lei and Yin, Zhenfei},
  journal={arXiv preprint arXiv:2506.09049},
  year={2025}
}

@inproceedings{grauman2022ego4d,
  title={Ego4d: Around the world in 3,000 hours of egocentric video},
  author={Grauman, Kristen and Westbury, Andrew and Byrne, Eugene and Chavis, Zachary and Furnari, Antonino and Girdhar, Rohit and Hamburger, Jackson and Jiang, Hao and Liu, Miao and Liu, Xingyu and others},
  booktitle={Proceedings of the IEEE/CVF conference on computer vision and pattern recognition},
  pages={18995--19012},
  year={2022}
}

@article{qin2025robofactory,
  title={Robofactory: Exploring embodied agent collaboration with compositional constraints},
  author={Qin, Yiran and Kang, Li and Song, Xiufeng and Yin, Zhenfei and Liu, Xiaohong and Liu, Xihui and Zhang, Ruimao and Bai, Lei},
  journal={arXiv preprint arXiv:2503.16408},
  year={2025}
}

@article{liu2025language,
  title={Language-driven policy distillation for cooperative driving in multi-agent reinforcement learning},
  author={Liu, Jiaqi and Xu, Chengkai and Hang, Peng and Sun, Jian and Ding, Mingyu and Zhan, Wei and Tomizuka, Masayoshi},
  journal={IEEE Robotics and Automation Letters},
  year={2025},
  publisher={IEEE}
}

@article{obata2024lip,
  title={LiP-LLM: Integrating Linear Programming and dependency graph with Large Language Models for multi-robot task planning},
  author={Obata, Kazuma and Aoki, Tatsuya and Horii, Takato and Taniguchi, Tadahiro and Nagai, Takayuki},
  journal={IEEE Robotics and Automation Letters},
  year={2024},
  publisher={IEEE}
}

@article{wang2024dart,
  title={Dart-llm: Dependency-aware multi-robot task decomposition and execution using large language models},
  author={Wang, Yongdong and Xiao, Runze and Kasahara, Jun Younes Louhi and Yajima, Ryosuke and Nagatani, Keiji and Yamashita, Atsushi and Asama, Hajime},
  journal={arXiv preprint arXiv:2411.09022},
  year={2024}
}

@article{zhang2023building,
  title={Building cooperative embodied agents modularly with large language models},
  author={Zhang, Hongxin and Du, Weihua and Shan, Jiaming and Zhou, Qinhong and Du, Yilun and Tenenbaum, Joshua B and Shu, Tianmin and Gan, Chuang},
  journal={arXiv preprint arXiv:2307.02485},
  year={2023}
}

@article{wang2025multi,
  title={Multi-Agent LLM Actor-Critic Framework for Social Robot Navigation},
  author={Wang, Weizheng and Obi, Ike and Min, Byung-Cheol},
  journal={arXiv preprint arXiv:2503.09758},
  year={2025}
}

@article{nasiriany2024robocasa,
  title={Robocasa: Large-scale simulation of everyday tasks for generalist robots},
  author={Nasiriany, Soroush and Maddukuri, Abhiram and Zhang, Lance and Parikh, Adeet and Lo, Aaron and Joshi, Abhishek and Mandlekar, Ajay and Zhu, Yuke},
  journal={arXiv preprint arXiv:2406.02523},
  year={2024}
}

@article{zhou2023sotopia,
  title={Sotopia: Interactive evaluation for social intelligence in language agents},
  author={Zhou, Xuhui and Zhu, Hao and Mathur, Leena and Zhang, Ruohong and Yu, Haofei and Qi, Zhengyang and Morency, Louis-Philippe and Bisk, Yonatan and Fried, Daniel and Neubig, Graham and others},
  journal={arXiv preprint arXiv:2310.11667},
  year={2023}
}

@article{bo2024reflective,
  title={Reflective multi-agent collaboration based on large language models},
  author={Bo, Xiaohe and Zhang, Zeyu and Dai, Quanyu and Feng, Xueyang and Wang, Lei and Li, Rui and Chen, Xu and Wen, Ji-Rong},
  journal={Advances in Neural Information Processing Systems},
  volume={37},
  pages={138595--138631},
  year={2024}
}

@article{guo2024embodied,
  title={Embodied llm agents learn to cooperate in organized teams},
  author={Guo, Xudong and Huang, Kaixuan and Liu, Jiale and Fan, Wenhui and V{\'e}lez, Natalia and Wu, Qingyun and Wang, Huazheng and Griffiths, Thomas L and Wang, Mengdi},
  journal={arXiv preprint arXiv:2403.12482},
  year={2024}
}

@article{zhang2024combo,
  title={COMBO: compositional world models for embodied multi-agent cooperation},
  author={Zhang, Hongxin and Wang, Zeyuan and Lyu, Qiushi and Zhang, Zheyuan and Chen, Sunli and Shu, Tianmin and Dariush, Behzad and Lee, Kwonjoon and Du, Yilun and Gan, Chuang},
  journal={arXiv preprint arXiv:2404.10775},
  year={2024}
}

@article{wang2025rad,
  title={RAD: Retrieval-Augmented Decision-Making of Meta-Actions with Vision-Language Models in Autonomous Driving},
  author={Wang, Yujin and Liu, Quanfeng and Jiang, Zhengxin and Wang, Tianyi and Jiao, Junfeng and Chu, Hongqing and Gao, Bingzhao and Chen, Hong},
  journal={arXiv preprint arXiv:2503.13861},
  year={2025}
}

@article{hu2023look,
  title={Look before you leap: Unveiling the power of gpt-4v in robotic vision-language planning},
  author={Hu, Yingdong and Lin, Fanqi and Zhang, Tong and Yi, Li and Gao, Yang},
  journal={arXiv preprint arXiv:2311.17842},
  year={2023}
}

@article{ji2025robobrain,
  title={Robobrain: A unified brain model for robotic manipulation from abstract to concrete},
  author={Ji, Yuheng and Tan, Huajie and Shi, Jiayu and Hao, Xiaoshuai and Zhang, Yuan and Zhang, Hengyuan and Wang, Pengwei and Zhao, Mengdi and Mu, Yao and An, Pengju and others},
  journal={arXiv preprint arXiv:2502.21257},
  year={2025}
}

@article{shi2024math,
  title={Math-llava: Bootstrapping mathematical reasoning for multimodal large language models},
  author={Shi, Wenhao and Hu, Zhiqiang and Bin, Yi and Liu, Junhua and Yang, Yang and Ng, See-Kiong and Bing, Lidong and Lee, Roy Ka-Wei},
  journal={arXiv preprint arXiv:2406.17294},
  year={2024}
}

@article{zhang2024mathematical,
  title={Mathematical visual instruction tuning with an automatic data engine},
  author={Zhang, Renrui and Wei, Xinyu and Jiang, Dongzhi and Guo, Ziyu and Li, Shicheng and Zhang, Yichi and Tong, Chengzhuo and Liu, Jiaming and Zhou, Aojun and Wei, Bin and others},
  journal={arXiv preprint arXiv:2407.08739},
  year={2024}
}

@article{gao2023g,
  title={G-llava: Solving geometric problem with multi-modal large language model},
  author={Gao, Jiahui and Pi, Renjie and Zhang, Jipeng and Ye, Jiacheng and Zhong, Wanjun and Wang, Yufei and Hong, Lanqing and Han, Jianhua and Xu, Hang and Li, Zhenguo and others},
  journal={arXiv preprint arXiv:2312.11370},
  year={2023}
}

@article{xu2411llava,
  title={Llava-cot: Let vision language models reason step-by-step, 2024},
  author={Xu, Guowei and Jin, Peng and Hao, Li and Song, Yibing and Sun, Lichao and Yuan, Li},
  journal={URL https://arxiv. org/abs/2411.10440},
  year={2024}
}

@article{wei2022chain,
  title={Chain-of-thought prompting elicits reasoning in large language models},
  author={Wei, Jason and Wang, Xuezhi and Schuurmans, Dale and Bosma, Maarten and Xia, Fei and Chi, Ed and Le, Quoc V and Zhou, Denny and others},
  journal={Advances in neural information processing systems},
  volume={35},
  pages={24824--24837},
  year={2022}
}

@article{guo2025deepseek,
  title={Deepseek-r1: Incentivizing reasoning capability in llms via reinforcement learning},
  author={Guo, Daya and Yang, Dejian and Zhang, Haowei and Song, Junxiao and Zhang, Ruoyu and Xu, Runxin and Zhu, Qihao and Ma, Shirong and Wang, Peiyi and Bi, Xiao and others},
  journal={arXiv preprint arXiv:2501.12948},
  year={2025}
}

@article{tan2025reason,
  title={Reason-rft: Reinforcement fine-tuning for visual reasoning},
  author={Tan, Huajie and Ji, Yuheng and Hao, Xiaoshuai and Lin, Minglan and Wang, Pengwei and Wang, Zhongyuan and Zhang, Shanghang},
  journal={arXiv preprint arXiv:2503.20752},
  year={2025}
}

@article{chen2025learning,
  title={Learning Only with Images: Visual Reinforcement Learning with Reasoning, Rendering, and Visual Feedback},
  author={Chen, Yang and Shen, Yufan and Huang, Wenxuan and Zhou, Sheng and Lin, Qunshu and Cai, Xinyu and Yu, Zhi and Bu, Jiajun and Shi, Botian and Qiao, Yu},
  journal={arXiv preprint arXiv:2507.20766},
  year={2025}
}

@inproceedings{jiang2024visual,
  title={Visual grounding for object-level generalization in reinforcement learning},
  author={Jiang, Haobin and Lu, Zongqing},
  booktitle={European Conference on Computer Vision},
  pages={55--72},
  year={2024},
  organization={Springer}
}

@article{zhang2024improve,
  title={Improve vision language model chain-of-thought reasoning},
  author={Zhang, Ruohong and Zhang, Bowen and Li, Yanghao and Zhang, Haotian and Sun, Zhiqing and Gan, Zhe and Yang, Yinfei and Pang, Ruoming and Yang, Yiming},
  journal={arXiv preprint arXiv:2410.16198},
  year={2024}
}

@inproceedings{zhu2017target,
  title={Target-driven visual navigation in indoor scenes using deep reinforcement learning},
  author={Zhu, Yuke and Mottaghi, Roozbeh and Kolve, Eric and Lim, Joseph J and Gupta, Abhinav and Fei-Fei, Li and Farhadi, Ali},
  booktitle={2017 IEEE International Conference on Robotics and Automation (ICRA)},
  pages={3357--3364},
  year={2017},
  organization={IEEE}
}

@article{zhou2025roborefer,
  title={RoboRefer: Towards Spatial Referring with Reasoning in Vision-Language Models for Robotics},
  author={Zhou, Enshen and An, Jingkun and Chi, Cheng and Han, Yi and Rong, Shanyu and Zhang, Chi and Wang, Pengwei and Wang, Zhongyuan and Huang, Tiejun and Sheng, Lu and others},
  journal={arXiv preprint arXiv:2506.04308},
  year={2025}
}

@article{sarch2025grounded,
  title={Grounded Reinforcement Learning for Visual Reasoning},
  author={Sarch, Gabriel and Saha, Snigdha and Khandelwal, Naitik and Jain, Ayush and Tarr, Michael J and Kumar, Aviral and Fragkiadaki, Katerina},
  journal={arXiv preprint arXiv:2505.23678},
  year={2025}
}

@article{liu2025noisyrollout,
  title={NoisyRollout: Reinforcing Visual Reasoning with Data Augmentation},
  author={Liu, Xiangyan and Ni, Jinjie and Wu, Zijian and Du, Chao and Dou, Longxu and Wang, Haonan and Pang, Tianyu and Shieh, Michael Qizhe},
  journal={arXiv preprint arXiv:2504.13055},
  year={2025}
}

@article{liu2025visual,
  title={Visual-rft: Visual reinforcement fine-tuning},
  author={Liu, Ziyu and Sun, Zeyi and Zang, Yuhang and Dong, Xiaoyi and Cao, Yuhang and Duan, Haodong and Lin, Dahua and Wang, Jiaqi},
  journal={arXiv preprint arXiv:2503.01785},
  year={2025}
}

@article{bai2025qwen2,
  title={Qwen2. 5-vl technical report},
  author={Bai, Shuai and Chen, Keqin and Liu, Xuejing and Wang, Jialin and Ge, Wenbin and Song, Sibo and Dang, Kai and Wang, Peng and Wang, Shijie and Tang, Jun and others},
  journal={arXiv preprint arXiv:2502.13923},
  year={2025}
}

@article{tao2024maniskill3,
  title={Maniskill3: Gpu parallelized robotics simulation and rendering for generalizable embodied ai},
  author={Tao, Stone and Xiang, Fanbo and Shukla, Arth and Qin, Yuzhe and Hinrichsen, Xander and Yuan, Xiaodi and Bao, Chen and Lin, Xinsong and Liu, Yulin and Chan, Tse-kai and others},
  journal={arXiv preprint arXiv:2410.00425},
  year={2024}
}

@misc{yin2025spatialmentalmodelinglimited,
      title={Spatial Mental Modeling from Limited Views}, 
      author={Baiqiao Yin and Qineng Wang and Pingyue Zhang and Jianshu Zhang and Kangrui Wang and Zihan Wang and Jieyu Zhang and Keshigeyan Chandrasegaran and Han Liu and Ranjay Krishna and Saining Xie and Manling Li and Jiajun Wu and Li Fei-Fei},
      year={2025},
      eprint={2506.21458},
      archivePrefix={arXiv},
      primaryClass={cs.AI},
      url={https://arxiv.org/abs/2506.21458}, 
}

@article{yuan2024robopoint,
  title={RoboPoint: A Vision-Language Model for Spatial Affordance Prediction for Robotics},
  author={Yuan, Wentao and Duan, Jiafei and Blukis, Valts and Pumacay, Wilbert and Krishna, Ranjay and Murali, Adithyavairavan and Mousavian, Arsalan and Fox, Dieter},
  journal={arXiv preprint arXiv:2406.10721},
  year={2024}
}

@inproceedings{10.1145/1121241.1121286,
  author = {Forlizzi, Jodi and DiSalvo, Carl},
  title = {Service robots in the domestic environment: a study of the roomba vacuum in the home},
  year = {2006},
  publisher = {Association for Computing Machinery},
  doi = {10.1145/1121241.1121286},
  booktitle = {Proceedings of the 1st ACM SIGCHI/SIGART Conference on Human-Robot Interaction},
  pages = {258--265},
  series = {HRI '06}
}

@misc{liu2023vehicletoeverythingautonomousdrivingsurvey,
  title={Towards Vehicle-to-everything Autonomous Driving: A Survey on Collaborative Perception},
  author={Si Liu and Chen Gao and Yuan Chen and Xingyu Peng and Xianghao Kong and Kun Wang and Runsheng Xu and Wentao Jiang and Hao Xiang and Jiaqi Ma and Miao Wang},
  year={2023},
  eprint={2308.16714},
  archivePrefix={arXiv},
  primaryClass={cs.CV}
}

@article{ye2025vlar1,
  title={{VLA-R1}: Enhancing Reasoning in Vision-Language-Action Models},
  author={Ye, Jiafei and others},
  journal={arXiv preprint arXiv:2505.xxxxx},
  year={2025}
}

@article{shen2025vlmr1,
  title={{VLM-R1}: A Stable and Generalizable {R1}-style Large Vision-Language Model},
  author={Shen, Yongqi and others},
  journal={arXiv preprint arXiv:2504.xxxxx},
  year={2025}
}

@misc{cai2025rocket2steeringvisuomotorpolicy,
  title={ROCKET-2: Steering Visuomotor Policy via Cross-View Goal Alignment},
  author={Shaofei Cai and Zhancun Mu and Anji Liu and Yitao Liang},
  year={2025},
  eprint={2503.02505},
  archivePrefix={arXiv},
  primaryClass={cs.AI}
}

@article{foutter2024space,
  title={Space-LLaVA: a Vision-Language Model Adapted to Extraterrestrial Applications},
  author={Foutter, Matthew and Gammelli, Daniele and Kruger, Justin and Foss, Ethan and Bhoj, Praneet and Guffanti, Tommaso and D'Amico, Simone and Pavone, Marco},
  journal={arXiv preprint arXiv:2408.05924},
  year={2024}
}

@inproceedings{deitke2025molmo,
  title={Molmo and pixmo: Open weights and open data for state-of-the-art vision-language models},
  author={Deitke, Matt and Clark, Christopher and Lee, Sangho and Tripathi, Rohun and Yang, Yue and Park, Jae Sung and Salehi, Mohammadreza and Muennighoff, Niklas and Lo, Kyle and Soldaini, Luca and others},
  booktitle={Proceedings of the Computer Vision and Pattern Recognition Conference},
  pages={91--104},
  year={2025}
}
 \bibliographystyle{iclr2026_conference}

\appendix
\newpage
{
\centering
\Large
\vspace{0.5em}Supplementary Material \\
}

\section{Limitations and Future Work}
\label{sec:limitations}

Despite the promising results, several limitations point to directions for future research. First, our framework operates on static, synchronized snapshots from all agents; in practice, distributed robots may capture observations at different times or with varying latencies. Extending CoRL to handle asynchronous sensing and temporal reasoning over video streams is an important next step. Second, the current architecture adopts a centralized design in which a single VLM processes all views jointly. While effective for a small number of agents, this imposes communication overhead and a single point of failure; its scalability to scenarios involving a large fleet of agents warrants further investigation into more efficient, perhaps hierarchical or message-passing-based, fusion strategies. Third, our benchmark does not yet address dynamic scenes with moving objects or agents, which would require tracking and re-identification across time as well as views. Fourth, the current coordinate prediction operates in 2D image space; incorporating depth information for full 3D spatial grounding would broaden applicability to tasks requiring precise manipulation in three dimensions. Finally, while our data covers 50+ object categories, the long-tail distribution of real-world environments suggests that further work on domain-invariant representations and open-vocabulary grounding could strengthen generalization.

This work opens several exciting directions. Future research could explore decentralized architectures where agents learn to communicate condensed, relevant information rather than sharing raw visual data. Another compelling avenue is the extension of CoRL to long-horizon, multi-step collaborative tasks, moving beyond single-step perception and grasping towards complex manipulation and navigation. Ultimately, we believe these directions offer promising avenues for advancing VLM-driven embodied collaboration and the construction of coherent world models from distributed views.

\section{Other Related Work and Novelty}

Recent work has begun to explore cross-view reasoning, yet these efforts remain fundamentally different from our problem setting. ROCKET-2~\cite{cai2025rocket2steeringvisuomotorpolicy} focuses on cross-view goal alignment for a \emph{single} visuomotor agent, where human and agent viewpoints differ; its approach introduces auxiliary losses (e.g., visibility and consistency losses) to improve policy cloning, but does not address the integration of multiple ego-centric observations or collaborative inference across agents. MindCube~\cite{yin2025spatialmentalmodelinglimited} instead evaluates whether VLMs can form \emph{spatial mental models} of static scenes from a few partial snapshots, emphasizing cognitive mapping and perspective-taking in a non-embodied setting without action generation or multi-agent coordination. Other concurrent works~\cite{yuan2024robopoint,foutter2024space,deitke2025molmo} investigate single-image spatial grounding or 3D-aware perception, but assume a single camera and do not consider distributed viewpoint fusion or embodied multi-view decision-making.

In contrast, our work introduces a \textbf{new embodied multi-agent setting} in which multiple robots simultaneously collect heterogeneous ego-centric views, and a central VLM must perform \emph{collaborative spatial reasoning} by fusing these distributed observations. This setting captures challenges---such as complementary visibility, cross-view ambiguity, and viewpoint-dependent grounding---that do not arise in prior single-agent or single-view formulations. To support systematic study of this problem, we present \textbf{E2W-Bench}, the first benchmark designed for world-centric reasoning from distributed embodied viewpoints across counting, relational QA, and view-specific grasp prediction. Moreover, to effectively solve this challenging setting, we propose \textbf{CoRL}, a two-stage SFT$\rightarrow$RL framework equipped with the \textbf{Cross-View Spatial Reward} (CVSR), which explicitly shapes multi-view grounding, overlap reasoning, and final answer correctness. Together, E2W-Bench and CoRL define a principled paradigm for multi-agent, multi-view spatial reasoning in embodied environments---going beyond prior cross-view or spatial QA works and establishing a novel direction for collaborative embodied AI.

\section{Problem Setup \& Notation}

In this section, we formally define the problem setup and introduce the notation used throughout the paper. We focus on the collaborative spatial reasoning task within multi-agent embodied systems, where multiple agents contribute to the construction of a shared understanding of a scene from partial observations.

\subsection{Problem Setup}

We consider a system of $N$ agents, each represented by an individual agent $a_i$ for $i \in \{1, \dots, N\}$. Each agent $a_i$ perceives the environment through a limited, ego-centric viewpoint, denoted by $\mathcal{V}_i$. The environment $\mathcal{E}$ is a spatial scene consisting of various objects, denoted by $\mathcal{O} = \{ o_1, o_2, \dots, o_M \}$, where $M$ is the total number of objects in the scene. Each object $o_j$ has associated properties, such as position, orientation, and object class, which we denote by $\mathbf{p}_j = (x_j, y_j, z_j)$, $\mathbf{r}_j = (\theta_j, \phi_j)$, and $c_j$ respectively.

The goal of the system is to enable the agents to collaboratively reason about the scene, including the localization of objects, the spatial relationships between objects, and the execution of spatially grounded actions. To achieve this, each agent $a_i$ shares its ego-centric observation $\mathcal{V}_i$ with a central Vision--Language Model (VLM), which integrates the individual observations to build a global understanding of the scene.

The task is formalized as a spatial reasoning problem, where the agents must complete a set of spatial tasks, such as object localization, relationship inference, and action planning, by collectively reasoning over their partial observations. Specifically, we aim to develop a system where the agents' observations are integrated effectively to:

\begin{itemize}
    \item Localize objects across different viewpoints.
    \item Infer spatial relationships between objects.
    \item Plan and execute actions based on a shared understanding of the scene.
\end{itemize}

\subsection{Notation}

We define the following notation used in this paper:

\begin{itemize}
    \item $\mathcal{A}$: Set of agents, where $\mathcal{A} = \{a_1, a_2, \dots, a_N\}$, and each agent $a_i$ has its own ego-centric observation $\mathcal{V}_i$.
    \item $\mathcal{O}$: Set of objects in the scene, $\mathcal{O} = \{o_1, o_2, \dots, o_M\}$.
    \item $\mathcal{V}_i$: Ego-centric observation of agent $a_i$, containing visual and linguistic information relevant to the agent's current view of the scene.
    \item $\mathbf{p}_j$: Position of object $o_j$, represented as a 3D vector $(x_j, y_j, z_j)$.
    \item $\mathbf{r}_j$: Orientation of object $o_j$, represented by angles $(\theta_j, \phi_j)$.
    \item $c_j$: Class of object $o_j$, which can be one of a predefined set of object categories.
    \item $Q$: A natural-language query describing the task to be performed.
    \item $\pi_\theta$: The central VLM policy parameterized by $\theta$, which integrates all ego-centric observations and the query to produce a prediction $\hat{y}$.
    \item $\hat{y}$: The model prediction, either a textual response (for counting and relational reasoning) or a pair of image-plane coordinates (for grasping).
    \item $R$: The reward function evaluating the quality of $\hat{y}$ against the ground-truth label $y$.
\end{itemize}

\subsection{Task Definition}

We define the collaborative spatial reasoning task as a sequential decision-making process where each agent must select an action from the set $\mathcal{A}_i$ based on its perception $\mathcal{V}_i$ and the shared scene information. The goal is to maximize the collective performance across all agents, measured by a global reward function that takes into account the accuracy of object localization, the correct inference of spatial relationships, and the successful execution of grounded actions.

The multi-agent system operates under the assumption of partial observability, meaning that each agent only has access to a limited subset of the scene and must rely on the collaboration of other agents to complete the task. The system must overcome the challenges of occlusions, incomplete observations, and ambiguous spatial relationships to successfully execute actions in a dynamic environment.

\subsection{Assumptions}

The following assumptions are made for the proposed problem setup:

\begin{itemize}
    \item Each agent has access to an ego-centric camera or sensor that provides partial observations of the scene.
    \item Agents can communicate with each other to share observations and jointly reason about the spatial layout of the environment.
    \item The central VLM has access to all agent observations and coordinates the integration of these inputs for collaborative reasoning.
    \item The agents' actions are executed in discrete time steps, and the environment responds to each action with a new state.
\end{itemize}

\section{E2W Benchmark Details}
\label{app:E2W}
In this section, we provide detailed information about the E2W data, which consists of both simulated and real-world data. The dataset includes over 100k simulated samples and more than 60k real-world samples. The benchmark is organized into three tasks, each with a distinct focus on collaborative spatial reasoning. Below, we describe the structure and data distribution for each task.

\subsection{Dataset Overview}

The E2W benchmark consists of the following components:

\begin{itemize}
    \item \textbf{Simulated Data}: A total of over 100k samples were collected from simulations, which provide a diverse set of scenes and spatial configurations for training and evaluation.
    \item \textbf{Real-World Data}: Over 60k samples were gathered from real-world environments, capturing the complexity of physical spaces and sensor noise that is typical in practical scenarios.
\end{itemize}

\subsection{Task Breakdown}

The E2W benchmark includes three distinct tasks designed to evaluate different aspects of collaborative spatial reasoning:

\begin{itemize}
    \item \textbf{Task 1 (Counting Task)}: This task focuses on the ability of the agents to count the number of objects within a scene. It is solely based on the simulated data, as real-world data for this task was not available.
    \item \textbf{Task 2 (Location Reasoning)}: This task evaluates the agents' ability to infer spatial relations among objects that never co-occur in a single view, requiring cross-view reasoning to answer correctly in natural language. Both simulated and real-world data are used, with the real-world data providing additional complexity due to sensor noise and occlusions.
    \item \textbf{Task 3 (Grasping)}: In this task, agents are required to translate a language command involving spatial relations into precise 2D coordinates within a specific agent's viewpoint, linking compositional reasoning to downstream robotic manipulation. Like Task 2, this task uses both simulated and real-world data.
\end{itemize}

\subsection{Dataset Splits}

For each task, we carefully curated the dataset as follows:

\begin{itemize}
    \item \textbf{Test Set}: For each task, we selected 200 high-quality samples to form the test set, ensuring that it covers a wide range of challenging scenarios.
    \item \textbf{Cold-Start Set (COT)}: We prepared a cold-start set consisting of 1000 samples, which is used to initialize the model before training. This set includes diverse configurations that enable the agents to begin learning without prior knowledge of the environment.
    \item \textbf{Training Set}: The remaining samples were used for training, with 90k samples from the simulated data contributing to the training set. The real-world data (6k samples) was integrated into the training set, but it is more sparsely used compared to the simulated data, ensuring a balance between generalizability and real-world applicability.
\end{itemize}

\begin{table}[!htbp]
    \centering
    \begin{tabular}{|c|c|c|c|}
    \hline
    \textbf{Task} & \textbf{Simulated Data} & \textbf{Real-World Data} & \textbf{Total Data} \\
    \hline
    Task 1 (Counting) & 30k & 0 & 30k \\
    Task 2 (Location Reasoning) & 35k & 25k & 60k \\
    Task 3 (Grasping) & 35k & 35k & 70k \\
    \hline
    \end{tabular}
    \caption{E2W Benchmark Dataset Distribution for Each Task}
    \label{tab:data_distribution}
\end{table}

\subsection{Data Quality and Selection Criteria}

Each task in the benchmark was carefully designed to cover a range of real-world challenges in collaborative spatial reasoning. To ensure high data quality, we followed strict selection criteria for both simulated and real-world data:

\begin{itemize}
    \item The simulated data was generated using a variety of scene configurations, object types, and spatial relationships to create a comprehensive and diverse training set.
    \item For the real-world data, we selected scenarios with clear object localization, minimal occlusions, and representative spatial relationships to maximize the relevance of the data for evaluating real-world performance.
    \item We prioritized edge cases and challenging scenarios for the test sets, ensuring that they push the limits of the agents' reasoning abilities and provide meaningful benchmarks for model performance.
\end{itemize}

\subsection{Data Collection}
\paragraph{Simulation Data} All task data are collected in a high-fidelity simulation environment built upon RoboFactory~\cite{qin2025robofactory} and ManiSkill3~\cite{tao2024maniskill3}, following scalable compositional simulation principles~\cite{qinbuilding}, which provide diverse scenes like RoboCasa~\cite{nasiriany2024robocasa}, articulated objects, and multi-agent configurations. We curate over 15,000 multi-agent samples across dozens of spatial layouts, each scene populated with a rich combination of manipulable objects under varied configurations. For every instance, synchronized global and egocentric camera views are recorded to support collaborative reasoning. Ground-truth annotations, including object counts, spatial relations, and manipulation-relevant attributes, are automatically derived from simulator metadata and physics engines, followed by human verification to ensure semantic consistency. This pipeline ensures that the resulting dataset captures the complementary demands of multi-view spatial reasoning, ranging from global aggregation to relational understanding and action grounding.

\paragraph{Real-World Data}
To complement simulation environments with naturalistic observations, we leverage the Ego4D dataset~\cite{grauman2022ego4d}, a massive-scale egocentric video corpus spanning 74 worldwide locations across 9 countries, with over 3,670 hours of daily-life recordings. Ego4D offers unconstrained visual contexts that reflect the challenges of embodied perception in realistic human environments. For data preparation, we first extract video frames at uniform intervals to obtain temporally diverse samples. We then apply strong vision backbones, including Region Attention Masking (RAM)~\cite{zhang2023recognize} for object-level proposals and DINO~\cite{caron2021emerging} for robust feature alignment, to automatically annotate object instances and spatial relations. Through this pipeline, we curate two subsets aligned with our benchmarks: \textbf{E2W-2-Real} (relational reasoning) and \textbf{E2W-3-Real} (grasping-oriented perception), each containing approximately 30k samples, yielding a total of 60k real-world instances.  This large-scale collection bridges the gap between simulation and reality by introducing natural visual noise, diverse object appearances, and unconstrained scene dynamics.

\section{Method Details (CoRL)}

In this section, we provide a detailed description of the proposed Collaborative Spatial Reasoning Reinforcement Learning (CoRL) framework. We outline the training objective, pipeline, and key components of the approach, which enable effective multi-agent collaborative spatial reasoning.

\subsection{Training Objective \& Pipeline}

The training objective for our proposed CoRL framework is designed to guide the agents towards improving their collaborative spatial reasoning capabilities. The main goal is to maximize the agents' ability to effectively integrate their partial, ego-centric observations and reason about the global spatial scene. To achieve this, we combine reinforcement learning (RL) with a novel Cross-View Spatial Reward (CVSR) function, which shapes the learning process by encouraging the agents to perform tasks that require accurate spatial localization, object relationships inference, and collaborative decision-making.

\subsubsection{Reinforcement Learning Setup}

We adopt a reinforcement learning setup, where each agent interacts with its environment and receives feedback in the form of a reward signal. The environment consists of a spatial scene with multiple agents, and each agent's objective is to complete a set of spatial reasoning tasks, such as object localization and relationship inference.

Each agent $a_i$ receives a partial observation $\mathcal{V}_i$ of the scene, which includes both visual and linguistic information. Based on this observation, the agent selects an action $a_i(t)$ from a predefined set of actions $y_i$. The agent's action affects the state of the environment, and the environment responds by providing a new state and a corresponding reward $r_i(t)$, which is calculated by the Cross-View Spatial Reward (CVSR) function.

The overall training objective is to maximize the expected cumulative reward for each agent:

\[
J(\theta) = \mathbb{E}\left[ \sum_{t=0}^{T} \gamma^t r_i(t) \right]
\]

where $T$ is the total number of time steps in the task, $\gamma$ is the discount factor, and $r_i(t)$ is the reward signal at time step $t$.

\subsection{Hyperparameters for Reward and CVSR}

In our CoRL framework, we define several hyperparameters for the reward function and Cross-View Spatial Reward (CVSR) that guide the agents' training process. These weights are carefully tuned to balance the different components of the task, ensuring that the agents effectively learn to perform spatial reasoning across multiple agents and viewpoints.

\subsubsection{CVSR Components Weights}

The Cross-View Spatial Reward (CVSR) function is designed to guide agents in overcoming challenges like occlusions, incomplete observations, and the need for integration across multiple viewpoints. The CVSR reward is composed of three components:

\[
R_{\mathrm{CVSR}}
= w_{\mathrm{ground}}\,R_{\mathrm{ground}}
+ w_{\mathrm{overlap}}\,R_{\mathrm{overlap}}
+ w_{\mathrm{ans}}\,R_{\mathrm{ans}},
\]

where the components are:

\begin{itemize}
    \item $w_{ans} = 0.7$: The weight for Answering (the ability to correctly complete the task based on spatial reasoning). This is the most important component, as it directly measures the success of the agent in solving the task.
    \item $w_{ground} = 0.1$: The weight for Localization (the ability to accurately localize objects in the scene). While important, this is a lower priority in comparison to answering tasks, as it is a fundamental skill that supports other reasoning tasks.
    \item $w_{overlap} = 0.2$: The weight for Fusion of Observations (the integration of ego-centric views). This component ensures the agents combine their partial observations to build a more complete understanding of the environment, and is essential for collaborative spatial reasoning.
\end{itemize}

These weights are chosen to emphasize the importance of answering correctly, while still ensuring that localization and fusion of observations are effectively learned.

\subsubsection{Training Pipeline}

The training pipeline for the CoRL framework follows a two-phase process: Supervised Fine-Tuning (SFT) followed by Reinforcement Learning (RL). The pipeline is as follows:

\begin{itemize}
    \item \textbf{Phase 1: Supervised Fine-Tuning (SFT)}: In this phase, the agents are first pre-trained using supervised learning. The goal is to initialize the model with basic spatial reasoning skills, using labeled data from the training set. The agents learn to perform tasks such as object localization and relationship inference based on ground truth labels.

    \item \textbf{Phase 2: Reinforcement Learning (RL)}: After the initial fine-tuning, the agents enter the RL phase, where they learn to improve their performance through interactions with the environment. The CVSR function is used to guide the agents' actions and refine their spatial reasoning capabilities. During this phase, agents iteratively adjust their policies to maximize the cumulative reward.
\end{itemize}

The overall training process is summarized in Algorithm \ref{alg:training_pipeline}.

\begin{algorithm}[htbp]
\caption{CoRL Training Pipeline}
\label{alg:training_pipeline}
\begin{algorithmic}[1]
\State \textbf{Input:} Pre-trained VLM parameters $\theta$, training dataset $\mathcal{D}$, CVSR weights $w_{\mathrm{ground}}, w_{\mathrm{overlap}}, w_{\mathrm{ans}}$, group size $G$
\State \textbf{Phase 1: Supervised Fine-Tuning (SFT)}
\For{each CoT-annotated sample $(x, q, r, a) \in \mathcal{D}$}
    \State Minimize $\mathcal{L}_{\mathrm{SFT}}$ (Eq.~3 in the main paper) via next-token prediction on $y = [r, a]$
\EndFor
\State Obtain SFT-initialized policy $\pi_0 \gets \pi_\theta$
\State \textbf{Phase 2: Reinforcement Learning (GRPO)}
\For{each training step}
    \State Sample input $u = (\{I_i\}_{i=1}^N, Q)$ from $\mathcal{D}$
    \State Generate $G$ candidate responses $\{y_j\}_{j=1}^G$ from $\pi_\theta$
    \State Score each $y_j$ with reward $R_j = \lambda_1 R_{\mathrm{format}} + \lambda_2 R_{\mathrm{CVSR}}$
    \State Compute group-relative advantages $\mathcal{A}_j = (R_j - \bar{R}) / \sigma_R$
    \State Update $\theta$ by maximizing $\mathcal{J}(\theta)$ (Eq.~7 in the main paper)
\EndFor
\end{algorithmic}
\end{algorithm}

\section{Real-World Robotic Evaluation}
We implement our real-world evaluation on manipulation platform equipped with two Franka Research~3 arms and a Realman mobile base. Each arm is paired with an Intel RealSense D435 RGB-D camera beside the robotic arm, and all cameras are extrinsically calibrated to a common base frame.

In the open-loop execution mode, CoRL-7B processes synchronized multi-view RGB images and predicts the 2D target location for the instructed action. The 2D coordinates are first fed into SAM2~\cite{ravi2024sam} to generate a segmentation mask, which filters the target object's point cloud from the RGB-D stream of the D435. The extracted point cloud is then passed to AnyGrasp~\cite{fang2023anygrasp}, which predicts a feasible grasp pose in the camera coordinate frame. Using the pre-calibrated camera extrinsics, the grasp pose is transformed into the coordinate system of the selected robotic arm. Then, we use the Deoxys library~\cite{zhu2022viola} to interact with the Franka Control Interface.

For placement, CoRL-7B outputs a 2D placement location, which is converted into 3D coordinates using the depth data. This 3D point is similarly transformed into the robot's coordinate frame, and the arm follows an open-loop trajectory to release the object at the designated position.

This setup allows us to directly connect vision--language inference with physical manipulation through SAM2-based segmentation and AnyGrasp-based grasp synthesis, without closed-loop corrections. It highlights the model's ability to produce actionable spatial predictions from multi-view observations.

\section{Prompt Design and Evaluation Protocols}
\label{sec:prompts}

\subsection{Spatial QA Generation}

\paragraph{Dual-Image Spatial Reasoning Prompt (with Auxiliary View).}
\begin{tcolorbox}[title=System Prompt, width=\textwidth, breakable]
\begin{Verbatim}[breaklines=true, breakanywhere=true, fontsize=\footnotesize]
You are provided with two images for multi-image reasoning. The first image shows labeled
objects with their bounding box coordinates: {object_labels}. The second image is an auxiliary
view taken {time_interval} seconds later that may provide additional context.

Please generate spatial relationship questions based primarily on the first image, while
having access to both images:
1. Generate 2-3 questions about the spatial relationship between the {obj1} and the {obj2}.
2. Base your questions on the labeled objects in the first image.
3. Use the coordinate information to understand object positions, but create questions and
   Chain-of-Thought (CoT) as if analyzing the first image.
4. Only ask questions relevant to their actual spatial relationship:
   - If objects are primarily horizontally separated: ask about left/right, not up/down/front/back.
   - If objects are primarily vertically separated: ask about up/down, not left/right/front/back.
5. Additional questions can be about distance, alignment, or blocking.
6. Use the actual object names in questions, not [A] [B] codes.
7. For each question, provide the correct answer (yes/no).
8. For each question, provide a CoT with two steps: first identify the bounding box coordinates
   from the first image, then briefly describe the spatial relationship.

Format your response as a JSON list of objects with 'question', 'answer', and 'cot' fields.
\end{Verbatim}
\end{tcolorbox}

\paragraph{Single-Image Spatial Reasoning Prompt.}
\begin{tcolorbox}[title=System Prompt, width=\textwidth, breakable]
\begin{Verbatim}[breaklines=true, breakanywhere=true, fontsize=\footnotesize]
You are looking at an image with labeled objects and their bounding box coordinates:
{object_labels}.

Please generate spatial relationship questions following these rules:
1. Generate 2-3 questions about the spatial relationship between the {obj1} and the {obj2}.
2. Use the coordinate information to understand object positions, but create questions and
   CoT as if analyzing a natural image.
3. Only ask questions relevant to their actual spatial relationship.
4. Use the actual object names in questions, not [A] [B] codes.
5. For each question, provide the correct answer (yes/no).
6. For each question, provide a CoT with two steps: first identify the bounding box
   coordinates, then briefly describe the spatial relationship.

Format your response as a JSON list of objects with 'question', 'answer', and 'cot' fields.
\end{Verbatim}
\end{tcolorbox}

\subsection{Where2Place Spatial Layout Prediction}

\begin{tcolorbox}[title=System Prompt, width=\textwidth, breakable]
\begin{Verbatim}[breaklines=true, breakanywhere=true, fontsize=\footnotesize]
You are an agent who reasons about free space and spatial relations. Based on the given image
and prompt, your task is to output a tuple (min_x, min_y, max_x, max_y) that represents the
bounding box of the object or region in the image that best matches the description in the
prompt. min_x, min_y, max_x, max_y should be integers indicating pixel locations. Do not
include any additional text in your answer other than the tuple (min_x, min_y, max_x, max_y).
\end{Verbatim}
\end{tcolorbox}

\begin{tcolorbox}[title=User Prompt, width=\textwidth, breakable]
\begin{Verbatim}[breaklines=true, breakanywhere=true, fontsize=\footnotesize]
Here is an image whose dimensions are labeled on the side. {question_text}
\end{Verbatim}
\end{tcolorbox}

\subsection{Multi-Robot View Analysis}

\begin{tcolorbox}[title=System Prompt, width=\textwidth, breakable]
\begin{Verbatim}[breaklines=true, breakanywhere=true, fontsize=\footnotesize]
You are analyzing two images from different robot perspectives viewing the same environment.
These images represent two distinct robot viewpoints that may show overlapping or complementary
parts of the scene. IMPORTANT FORMAT REQUIREMENTS:
1. ALL reasoning must be enclosed within <think> </think> tags.
2. When identifying objects RELEVANT TO THE QUESTION, provide their bounding box coordinates
   in the format: [x1, y1, x2, y2].
3. Only include bounding boxes for objects directly related to answering the question.
4. The final numerical answer MUST be placed in \boxed{NUMBER} - put ONLY the number inside
   the braces.
5. Do NOT put coordinates, lists, or other content inside \boxed{}.

EXAMPLE FORMAT:
<think>
I need to analyze both robot perspectives to count the requested objects.
In the main image, I can see a cup at bounding box [123, 150, 212, 250] and another cup at
[215, 100, 122, 180].
In the auxiliary image, I can see one cup at [150, 200, 250, 311].
Combining both perspectives, I count 3 cups total.
</think>

\boxed{3}
\end{Verbatim}
\end{tcolorbox}

\paragraph{User Message Format.}
\begin{tcolorbox}[title=User Message, width=\textwidth, breakable]
\begin{Verbatim}[breaklines=true, breakanywhere=true, fontsize=\footnotesize]
main picture: <image>
auxiliary picture: <image>
Question: {object_counting_question}
\end{Verbatim}
\end{tcolorbox}

\section{Case Studies}
\begin{figure}[H]
    \centering
    \includegraphics[width=0.9\linewidth]{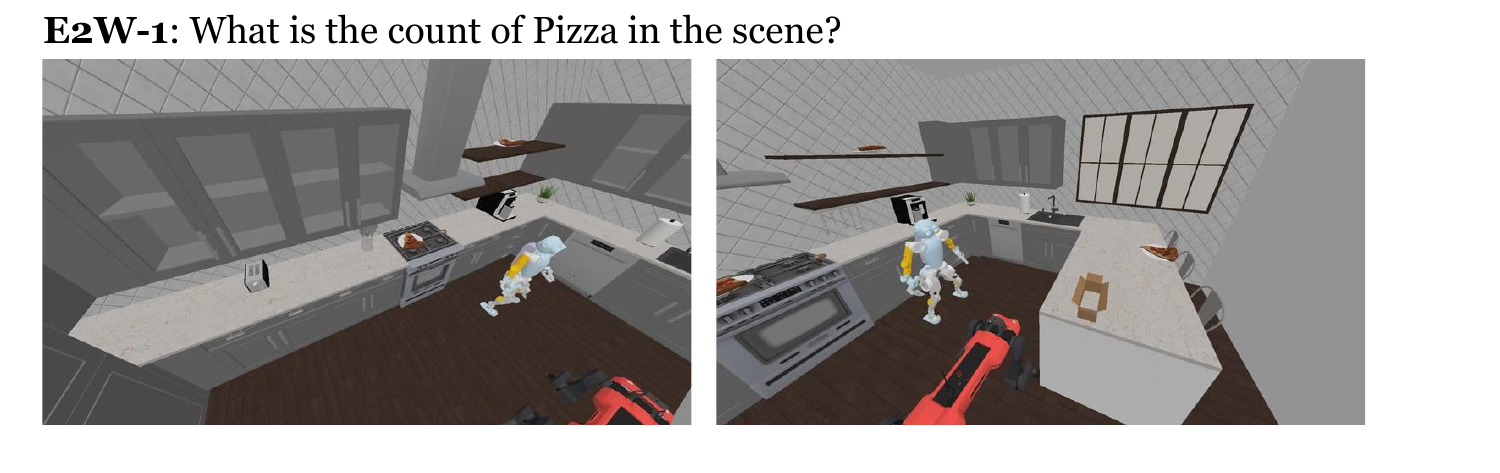}
    \caption{Example of an E2W-1 sample (Case 1). Two ego-centric views of a kitchen scene are provided, and the model must count the total number of pizzas by resolving cross-view object correspondences.}
    \label{fig:case_pizza}
\end{figure}
% -------- Model Responses --------
\begin{tcolorbox}[title=Model Responses (Case 1: Pizza Counting), width=\textwidth, breakable]
\textbf{Ground truth answer:} \(\boxed{3}\)

\medskip
\textbf{Our model (correct):}
\begin{Verbatim}[breaklines=true]
Our model: <think>\nI need to analyze images from two robot perspectives to answer the question about 'How many pizza do you observe?'.\n\nLet me first examine the main perspective (Image A):\n- I can see the first pizza at [419, 204, 451, 224]\n- I also see a second pizza at [532, 94, 571, 102]\n\nNow examining the auxiliary perspective (Image B):\n- In this view, I can see the first pizza at [0, 253, 44, 280]\n- I also notice a second pizza located at [499, 218, 544, 238]\n- Additionally, there's a third pizza at [161, 101, 200, 110]\n\nAnalyzing spatial correspondence between the two perspectives:\n- pizza_0 visible in main perspective at [419, 204, 451, 224] corresponds to the same object visible in auxiliary perspective at [0, 253, 44, 280]\n- pizza_1 at [499, 218, 544, 238] in auxiliary perspective is not visible in main perspective\n- pizza_2 visible in main perspective at [532, 94, 571, 102] corresponds to the same object visible in auxiliary perspective at [161, 101, 200, 110]\n\nCombining observations from both perspectives, the total number of objects you asked me to count is 3.\n</think>\n\n\\boxed{3}
\end{Verbatim}

\medskip
\textbf{GPT-5 (incorrect, predicts 2):}
\begin{Verbatim}[breaklines=true]
Gpt-5: I can see two pieces of pizza:
One slice on the upper wooden shelf (visible in both images).
One slice on the counter near the barstools.
The item on the stove looks like a roasted chicken or meat, not pizza.
So the answer is: 2 pizzas (slices).
\end{Verbatim}

\medskip
\textbf{Qwen-max (incorrect, predicts 2):}
\begin{Verbatim}[breaklines=true]
Qwen-max: Based on the two images provided, I can observe a total of two pizzas.
Here is the breakdown:
One pizza is on the stovetop in the first image and remains there in the second image.
A second pizza is visible on the kitchen island in the second image. It appears to be sitting next to an open cardboard box, likely a pizza box.
Therefore, across both views of the kitchen, there are 2 pizzas visible.
\end{Verbatim}
\end{tcolorbox}

\begin{figure}[H]
    \centering
    \includegraphics[width=0.9\linewidth]{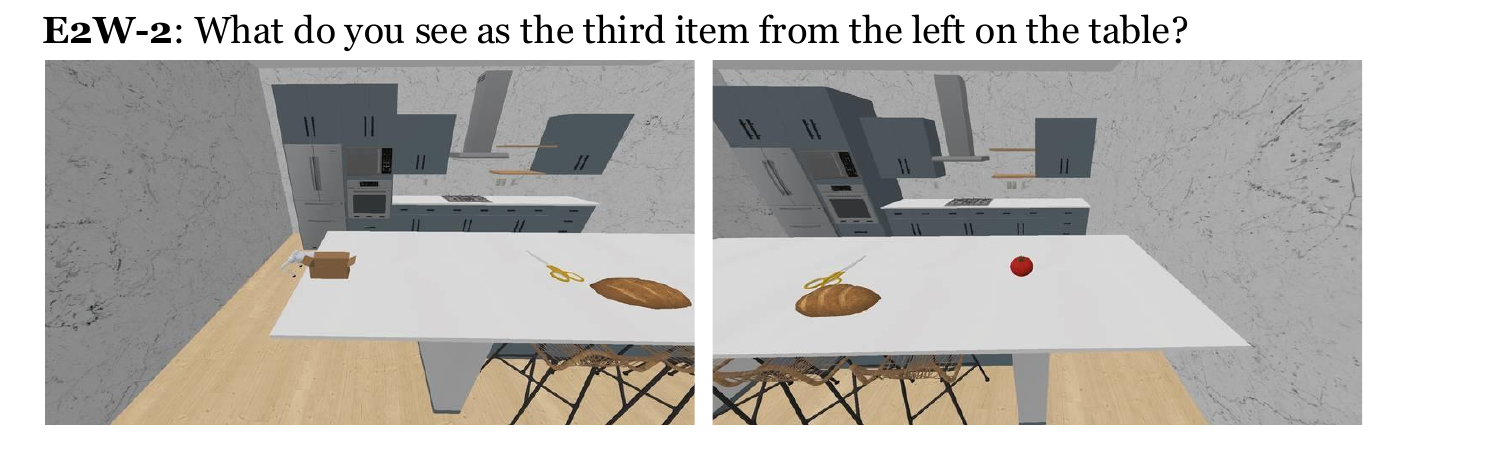}
    \caption{Example of an E2W-2 sample (Case 2). The model receives two views of a table scene and must identify the third item from the left by integrating spatial information across views.}
    \label{fig:case_table}
\end{figure}

% -------- Model Responses --------
\begin{tcolorbox}[title=Model Responses (Case 2: Third Item on Table), width=\textwidth, breakable]
\textbf{Ground truth answer:} \(\boxed{\text{bread}}\)

\medskip
\textbf{Our model (correct):}
\begin{Verbatim}[breaklines=true]
Our model: <think>\nI need to analyze images from two robot perspectives to answer the spatial reasoning question: 'What do you see as the third item from the left on the table?'.\n\nLet me first examine the main perspective (Image A):\n- I can see a cardboardbox located at [150, 113, 183, 129]\n- There's a scissors positioned at coordinates [283, 112, 318, 130]\n- I notice a bread at [321, 128, 381, 147]\n\nSpatial relationships in the main perspective:\n- cardboardbox is to the left of scissors (distance: 134.0 pixels)\n- cardboardbox is to the left of bread (distance: 185.2 pixels)\n- scissors is to the left of bread (distance: 53.1 pixels)\n\nNow examining the auxiliary perspective (Image B):\n- I can see a scissors located at [54, 115, 90, 135]\n- There's a bread positioned at coordinates [48, 132, 99, 152]\n- I notice a tomato at [175, 115, 188, 127]\n\nSpatial relationships in the auxiliary perspective:\n- scissors is near bread (distance: 17.1 pixels)\n- scissors is to the left of tomato (distance: 109.6 pixels)\n- bread is to the left of tomato (distance: 110.0 pixels)\n\nAnalyzing spatial correspondence between the two perspectives:\n- cardboardbox is only visible in main perspective at [150, 113, 183, 129]\n- scissors is visible in both perspectives at [283, 112, 318, 130] and [54, 115, 90, 135] respectively\n- bread is visible in both perspectives at [321, 128, 381, 147] and [48, 132, 99, 152] respectively\n- tomato is only visible in auxiliary perspective at [175, 115, 188, 127]\n\nTherefore, the answer to 'What do you see as the third item from the left on the table?' is bread.\n</think>\n\n\\boxed{bread}
\end{Verbatim}

\medskip
\textbf{GPT-5 (incorrect, predicts tomato):}
\begin{Verbatim}[breaklines=true]
Gpt-5: On the table, from left to right in the images, I see:
A loaf of bread
A pair of scissors
A tomato
So the third item from the left is a tomato.
\end{Verbatim}

\medskip
\textbf{Qwen-max (incorrect, focuses on tomato in second image):}
\begin{Verbatim}[breaklines=true]
Qwen-max: Based on the two images provided, the items on the table appear to have been rearranged.
Let's analyze each image:
In the first image: Looking from left to right on the white table, we can identify:
A small brown cardboard box.
A pair of yellow-handled scissors.
A loaf of bread.
Therefore, in the first image, the third item from the left is the loaf of bread.
In the second image: The arrangement has changed. From left to right, we see:
The loaf of bread.
The yellow-handled scissors.
A single red tomato in the center of the table.
Therefore, in the second image, the third item from the left is the tomato.
Since you asked "what do you see," and presented two different views, it's important to note the difference. However, if we consider the most complete or final view, the second image shows a more organized setup with the tomato placed centrally.
The third item from the left on the table in the second image is a tomato.
\end{Verbatim}
\end{tcolorbox}

\begin{figure}[H]
    \centering
    \includegraphics[width=0.9\linewidth]{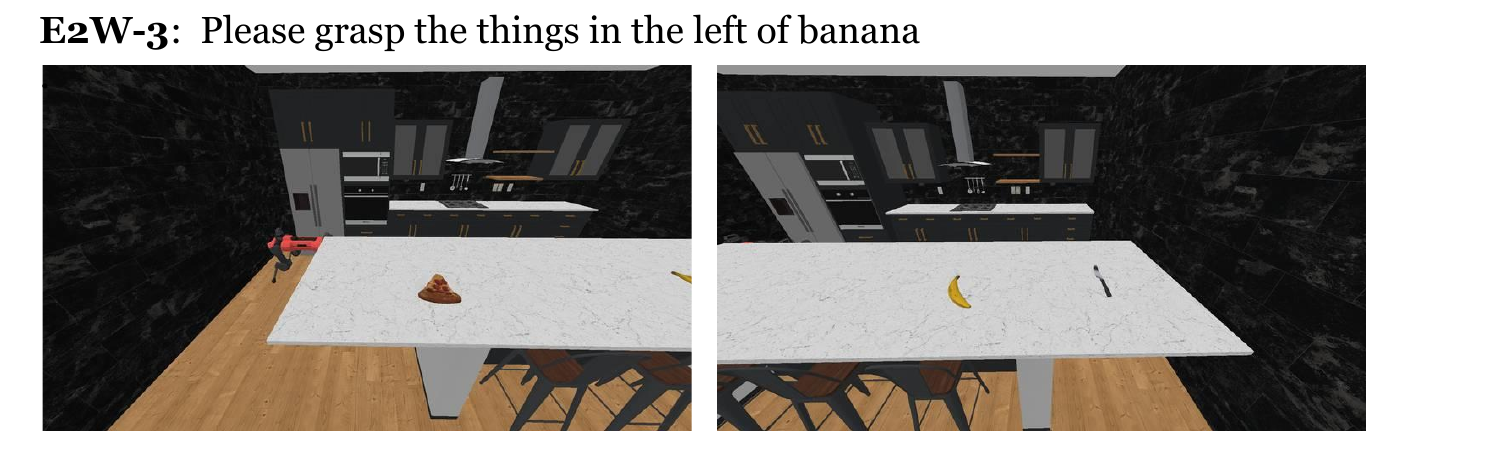}
    \caption{Example of an E2W-3 sample. The model receives two ego-centric views from different robot perspectives and must localize the target object---in this case, the pizza---and output the precise grasp location. Solving the task requires cross-view object correspondence, reasoning about spatial ordering, identifying the visible target in each view, and determining the final grasp coordinate in the main perspective.}
    \label{fig:case_e2w3}
\end{figure}

\begin{tcolorbox}[title=Model Responses (Case 3: E2W-3 Grasp Localization), width=\textwidth, breakable]

\medskip
\textbf{Our model:}
\begin{Verbatim}[breaklines=true]
Our model: <think>\nI need to analyze images from two robot perspectives to determine
the precise grasping location for the instruction:
'Please grasp the things in the left of banana.'.\n\n

Let me first examine the main perspective (Image A):\n
- I can see a banana located at [742, 243, 767, 258] with center at [754.5, 250.5]\n
- There's a pizza positioned at coordinates [440, 246, 495, 281]
  with center at [467.5, 263.5]\n\n

Spatial arrangement from left to right in main perspective:\n
- Position 1: pizza at center [467.5, 263.5]\n
- Position 2: banana at center [754.5, 250.5]\n\n

Now examining the auxiliary perspective (Image B):\n
- I can see a banana at [272, 248, 299, 286], center [285.5, 267.0]\n
- There's a knife at [445, 235, 466, 272], center [455.5, 253.5]\n\n

Overlap number = 1. Spatial ordering:\n
- Position 1: banana\n
- Position 2: knife\n\n

Analyzing target object:\n
The target is the pizza, which is left of the banana in the main perspective.\n\n

Pizza bounding box: [440, 246, 495, 281]\n
Center = [467.5, 263.5]\n\n

Thus I should grasp at image 0 coordinates [467.5, 263.5].\n
</think>\n\n
\\boxed{0, [467.5, 263.5]}
\end{Verbatim}

\end{tcolorbox}

\end{document}